\documentclass[pdflatex,sn-mathphys-num]{sn-jnl}

\usepackage{graphicx}%
\usepackage{multirow}%
\usepackage{amsmath,amssymb,amsfonts}%
\usepackage{amsthm}%
\usepackage{mathrsfs}%
\usepackage[title]{appendix}%
\usepackage{xcolor}%
\usepackage{textcomp}%
\usepackage{manyfoot}%
\usepackage{booktabs}%
\usepackage{algorithm}%
\usepackage{algorithmicx}%
\usepackage{algpseudocode}%
\usepackage{listings}%

\theoremstyle{thmstyleone}%
\theoremstyle{thmstyletwo}%

\theoremstyle{thmstylethree}%

\usepackage{orcidlink}
\newcommand{\methodname}{scSGC}
\usepackage{booktabs}
\usepackage{makecell}
\usepackage{caption}
\usepackage{overpic}
\graphicspath{{Fig/}}
\usepackage{multicol}
\usepackage{multirow} 
\usepackage{bm}
\usepackage[font=tiny]{subfig}

\begin{document}

\title[Article Title]{Soft Graph Clustering for single-cell RNA Sequencing Data}


\author[1,2]{\fnm{Ping} \sur{Xu}}\email{xuping@cnic.con}
\author[1,2]{\fnm{Pengfei} \sur{Wang}}\email{pfwang@cnic.cn}
\author[1,2]{\fnm{Zhiyuan} \sur{Ning}}\email{ningzhiyuan@cnic.cn}
\author[1,2]{\fnm{Meng} \sur{Xiao}}\email{shaow@cnic.cn}
\author[4]{\fnm{Min} \sur{Wu}}\email{wumin@i2r.a-star.edu.sg}
\author*[1,2,3]{\fnm{Yuanchun} \sur{Zhou}}\email{zyc@cnic.cn}

\affil*[1]{\orgdiv{Computer Network Information Center}, \orgname{Chinese Academy of Sciences}, \orgaddress{\postcode{100083}, \state{Beijing}, \country{China}}}

\affil[2]{\orgname{University of Chinese Academy of Sciences}, \orgaddress{\postcode{100864}, \state{Beijing}, \country{China}}}

\affil[3]{\orgdiv{Hangzhou Institute for Advanced Study}, \orgname{Chinese Academy of Sciences}, \orgaddress{\postcode{310024}, \state{Hangzhou}, \country{China}}}

\affil[4]{\orgdiv{Institute for Infocomm Research (I2R)}, \orgname{Agency for Science, Technology and Research (A*STAR)}, \orgaddress{\postcode{138632}, \country{Singapore}}}


\abstract{
\textbf{Background:} Clustering analysis is fundamental in single-cell RNA sequencing (scRNA-seq) data analysis for elucidating cellular heterogeneity and diversity. Recent graph-based scRNA-seq clustering methods, particularly graph neural networks (GNNs), have significantly improved in tackling the challenges of high-dimension, high-sparsity, and frequent dropout events that lead to ambiguous cell population boundaries. However, one major challenge for GNN-based methods is their reliance on hard graph constructions derived from similarity matrices. These constructions introduce difficulties when applied to scRNA-seq data due to:
(i) The simplification of intercellular relationships into binary edges (0 or 1) by applying thresholds, which restricts the capture of continuous similarity features among cells and leads to significant information loss.
(ii) The presence of significant inter-cluster connections within hard graphs, which can confuse GNN methods that rely heavily on graph structures, potentially causing erroneous message propagation and biased clustering outcomes.

\textbf{Results:} To tackle these challenges, we introduce~\textbf{\methodname}, a \textbf{S}oft \textbf{G}raph \textbf{C}lustering for \textbf{s}ingle-\textbf{c}ell RNA sequencing data, which aims to more accurately characterize continuous similarities among cells through non-binary edge weights, thereby mitigating the limitations of rigid data structures. 
The~\methodname~framework comprises three core components: 
(i) a zero-inflated negative binomial (ZINB)-based feature autoencoder designed to effectively handle the sparsity and dropout issues in scRNA-seq data; 
(ii) a dual-channel cut-informed soft graph embedding module, constructed through deep graph-cut information, capturing continuous similarities between cells while preserving the intrinsic data structures of scRNA-seq; 
and (iii) an optimal transport-based clustering optimization module, achieving optimal delineation of cell populations while maintaining high biological relevance. 

\textbf{Conclusion:} 
By integrating dual-channel cut-informed soft graph representation learning, a ZINB-based feature autoencoder, and optimal transport-driven clustering optimization, scSGC effectively overcomes the challenges associated with traditional hard graph constructions in GNN methods. Extensive experiments across ten datasets demonstrate that scSGC outperforms 13 state-of-the-art clustering models in clustering accuracy, cell type annotation, and computational efficiency. These results highlight its substantial potential to advance scRNA-seq data analysis and deepen our understanding of cellular heterogeneity.
}




\maketitle

\section{\textbf{Background}}
\label{sec:introduction}
Single-cell RNA sequencing (scRNA-seq) technology precisely captures gene expression differences at the single-cell level, revealing cellular heterogeneity and diversity, and providing crucial insights for bioinformatics~\cite{yang2024genecompass}. 
Clustering cells based on gene expression patterns is essential in scRNA-seq analysis, aiding in cell type annotation and marker gene identification~\cite{kiselev2019challenges, wang2025sccompass}. 
Accurately identifying pathological cell types through clustering provides important information for clinical diagnosis and personalized treatment, thereby significantly advancing precision medicine~\cite{haghverdi2016diffusion}.

Cell clustering has garnered significant research interest, resulting in the development of various methods such as K-means and hierarchical clustering. 
Advanced techniques like Phenograph~\cite{levine2015data}, MAGIC~\cite{van2018recovering}, and Seurat~\cite{butler2018integrating} employ k-nearest neighbor (KNN) algorithms to model cellular relationships, with MAGIC pioneering explicit genome-wide imputation in single-cell expression profiles. 
In contrast, CIDR~\cite{lin2017cidr} uses implicit imputation to mitigate the effects of dropout events. 
Multi-kernel spectral clustering approaches, including SIMLR~\cite{wang2018simlr} and MPSSC~\cite{park2018spectral}, utilize multiple kernel functions for robust similarity measure learning; MPSSC is noted for its effectiveness in handling scRNA-seq data sparsity. 
Despite these advancements, scRNA-seq data still face challenges like high-sparsity and frequent dropout events, where many genes are undetected or incorrectly detected, leading to blurred cell boundaries. 
Additionally, advances in sequencing technologies have increased data volume and dimensionality, which constrain existing methods' capability to accurately capture intercellular similarities in high-dimensional space~\cite{wang2025sccompass}.

With the growing focus on cellular heterogeneity, deep learning has become a central technique in scRNA-seq clustering. 
Techniques like DCA leverage zero-inflated negative binomial (ZINB) autoencoders to model scRNA-seq data distribution, 
while scDeepCluster~\cite{tian2019clustering} integrates ZINB-based autoencoders with deep-embedded clustering to optimize latent feature learning and improve clustering performance. 
AttentionAE-sc~\cite{li2023attention} employs attention mechanisms to combine denoised representation learning with clustering-friendly feature extraction. 
Despite these advancements, many methods overlook intercellular structural information~\cite{gan2018identification}. 
To address this, algorithms such as Louvain and Leiden~\cite{traag2019louvain} utilize graph structures to better capture cellular similarities. Approaches like scGAE~\cite{luo2021topology} and graph-sc~\cite{ciortan2022gnn} employ graph autoencoders to embed data while preserving topology. 
Similarly, scGNN~\cite{wang2021scgnn} merges graph neural networks (GNNs) with multimodal autoencoders to aggregate cellular relationships and model gene expression patterns using a left-truncated Gaussian mixture model. 
Likewise, scDSC~\cite{gan2022deep} integrates a ZINB-based autoencoder with a GNN module, using a mutual supervision strategy to optimize data representation.

Although GNN-based methods have advanced general graph encoding~\cite{zhou2024make},  they face specific challenges when applied to scRNA-seq data: 
(i) \textbf{Dependence on hard graph constructions}: GNN-based methods typically rely on hard graph constructions, where edges between nodes (cells) are defined using strict criteria or thresholds, i.e., edges are represented as binary (0/1) connections~\cite{ning2021lightcake,alekseev2007np,tang2021graph,wu2025soft}. 
This binary approach oversimplifies the complexity of cellular similarities by disregarding transitional states between cells.   
(ii) \textbf{Similar representations for nearby nodes}: GNN-based methods rely heavily on the underlying graph structure when processing data, aggregating information from neighboring nodes through inherent message-passing mechanisms to generate similar representations~\cite{ning2022graph, ning2025deep,chen2020measuring,ning2025rethinking}. 
However, hard graphs constructed from high-dimensional and high-sparse scRNA-seq data often exhibit significant inter-cluster connections, potentially resulting in erroneous information propagation between cells that should belong to distinct clusters. 

To tackle these challenges, we propose a \textbf{S}oft \textbf{G}raph \textbf{C}lustering for \textbf{s}ingle-\textbf{c}ell RNA sequencing data, namely \textbf{\methodname}, which aims to leverage soft graph construction to more accurately capture the continuous similarities between cells through non-binary edge weights. 
By overcoming the limitations of rigid binary structures,~\methodname~facilitates improved identification of distinct cellular subtypes and clearer delineation of cell populations, , thereby advancing our understanding of cellular heterogeneity. 
In~\methodname, we first utilize a ZINB autoencoder to handle the sparsity and dropout issues inherent in scRNA-seq data, generating robust cellular representations. 
Then, two soft graphs are constructed using the input data, and their corresponding laplacian matrices are computed~\cite{wu2025soft,thumbakara2014soft, wang2024comprehensive}. 
These matrices undergo a minimum jointly normalized cut through a graph-cut strategy to optimize the representation of cell-cell relationships~\cite{xu2024sccdcg}. 
Finally, an optimal transport-based self-supervised learning approach is employed to refine the clustering, ensuring accurate partitioning of cell populations in high-dimensional and high-sparse data~\cite{xu2025scsiameseclu}. 
Experimental results across ten datasets demonstrate that \textbf{\methodname} outperforms eleven state-of-the-art (SOTA) single-cell clustering models and exhibits competitive performance across various downstream tasks, highlighting its robustness and effectiveness in diverse analytical scenarios.

\section{\textbf{Methods}}

\begin{figure*}[t!]
    \centering
    \includegraphics[width=1\textwidth]{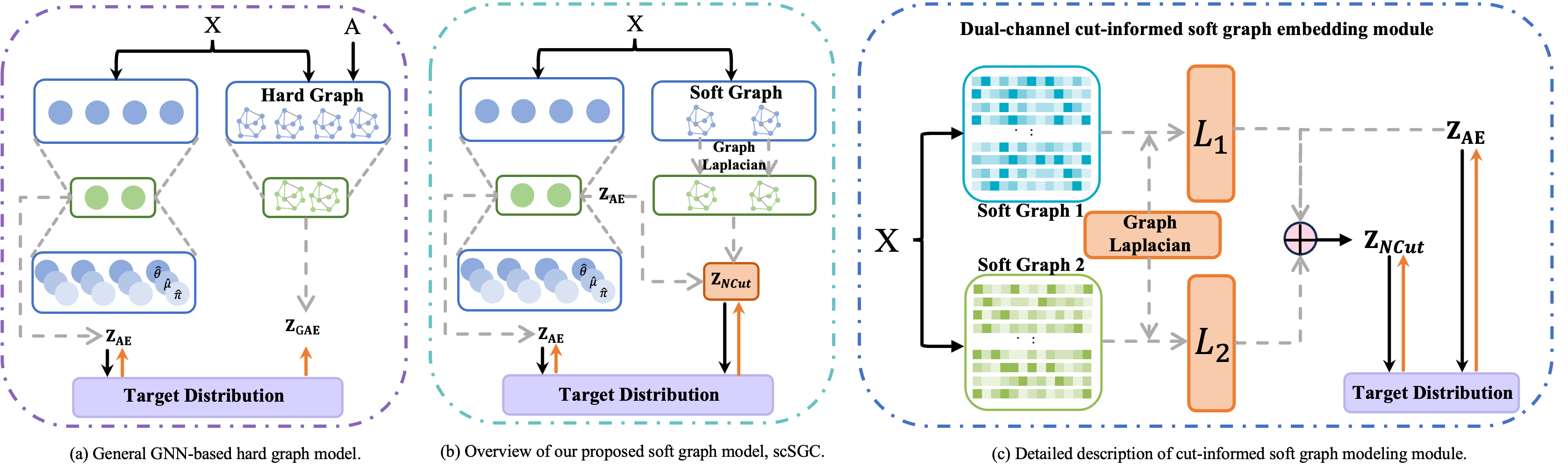}
    \caption{Comparison for diagram of graph-based scRNA-seq data clustering framework and detailed description of cut-informed soft graph modeling module. 
    }
    \label{fig:overview}
    \vspace{-4mm}
\end{figure*} 

\subsection{\textbf{The framework of~\methodname}}
Fig.~\ref{fig:overview}(a) depicts a hard graph GNN-based framework for scRNA-seq clustering, while Fig.~\ref{fig:overview}(b) illustrates the framework of our proposed method,\textbf{~\methodname}.
Unlike traditional methods, \textbf{\methodname} provides two key advantages: 
(i) It tightly integrates two key modules within the graph-based scRNA-seq clustering framework, i.e., the feature autoencoder and the graph autoencoder, allowing both modules to optimize the final embedding collaboratively; 
(ii) By employing a soft graph construction strategy, it eliminates reliance on hard graph structures, enabling more effective capture of intracellular structural information and fully utilizing continuous similarities between cells.
Specifically, our proposed method,~\methodname, comprises three key modules: 
(i) \textbf{ZINB-based feature autoencoder}, which employs a ZINB model to characterize scRNA-seq data specifically to address high sparsity and dropout rates in gene expression data; 
(ii) \textbf{Cut-informed soft graph modeling} (see Fig.~\ref{fig:overview}(c) for its architecture), leverages dual-channel cut-informed soft graph construction to generate consistent and optimized cellular representations, facilitating smoother capture of intercellular continuous structural information; 
(iii) \textbf{Optimization via optimal transport}, utilizing optimal transport theory, achieving optimal partitioning of cell populations at minimal transport cost and ensuring stable clustering results within complex data structures. 

\subsection{\textbf{Methodology}}
To begin, denote 
the preprocessed input data as $\mathbf{X}$,
where $x_{ij}\left(1\leq i\leq N, 1\leq j\leq D\right)$ represents the expression value of $j$-th gene for the $i$-th cell in the dataset. 
In this study, we aim to simultaneously learn the feature representation $\mathbf{Z}$ for each cell and the clustering assignment $\mathbf{C}$.  

\subsubsection{\textbf{ZINB-based feature autoencoder}}
To accurately model the distribution of scRNA-seq data and address the challenges posed by its high sparsity and frequent dropout events, we introduce a ZINB-based autoencoder, which utilizes the log-likelihood function of the ZINB distribution as its loss function.
\begin{equation}
\mathrm{NB}(\mathbf{X}_{ij}; \mu_{ij}, \theta_{ij})=\frac{\Gamma(\mathbf{X}_{ij}+\theta_{ij})}{\Gamma(\theta_{ij})}\left(\frac{\theta_{ij}}{\theta_{ij}+\mu_{ij}}\right)^{\theta_{ij}}\left(\frac{\mu_{ij}}{\theta_{ij}+\mu_{ij}}\right)^{\mathbf{X}_{ij}},
\label{equ:nb}
\end{equation}
\begin{equation}
\operatorname{ZINB}(\mathbf{X}_{ij} ; \pi_{ij}, \mu_{ij}, \theta_{ij})=\pi_{ij} \delta_0(\mathbf{X}_{ij})+(1-\pi_{ij}) \mathrm{NB}(\mathbf{X}_{ij} ; \mu, \theta),
\label{equ:zinb}
\end{equation}
where dropout ($\pi_{ij}$) is the probability of dropout events, mean ($\mu_{ij}$) and dispersion ($\theta_{ij}$) parameters, respectively, represent the negative binomial component. 

Then, we define the encoder function as $\mathbf{Z}=f_e\left(\mathbf{X}\right)$ and the decoder function as $\mathbf{X}^{'}=f_d\left(\mathbf{Z}\right)$, where both the encoder and decoder are implemented as fully connected neural networks. 
The decoder has three output layers to estimate parameters $\pi$, $\mu$ and $\theta$, which are
\begin{equation}
\begin{aligned}
&\hat{\pi}=\operatorname{sigmoid}\left(W_\pi \cdot \mathbf{Z}\right), \\
&\hat{\mu}=diag\left(s_i\right)\times\exp \left(W_\mu \cdot \mathbf{Z}\right), \\
&\hat{\theta}=\exp \left(W_\theta \cdot \mathbf{Z}\right).
\end{aligned}
\label{equ:zinb_parameters}
\end{equation}
Similarly, $W_\pi$, $W_\mu$ and $W_\theta$ are the corresponding weights, the size factors $s_i$ is the ratio of the total cell count to the median $S$. 
Finally, we minimize the overall negative likelihood loss of the  ZINB model-based autoencoder, 
\begin{equation}
\mathcal{L}_{Z I N B }=-log\left(Z I N B\left(\mathbf{X}_{ij} \mid \hat{\pi}_{ij}, \hat{\mu}_{ij}, \hat{\theta}_{ij}\right)\right)
\label{loss_zinb}
\end{equation}

\subsubsection{\textbf{Cut-informed soft graph modeling}} 
We propose the dual-channel cut-informed soft graph embedding module to capture the continuous similarities between cells from high-dimensional and high-sparse scRNA-seq data. 
As depicted in Fig.~\ref{fig:overview}(c), this module constructs dual-channel soft graphs and computes the corresponding laplacian matrices, then applies a minimum jointly normalized cut strategy to fuse the matrices from both channels.

Initially, using the preprocessed scRNA-seq data matrix $\mathbf{X}$ as input, we adopt a dual-channel strategy to construct two distinct similarity soft graphs: the feature similarity graph $\mathcal{G}_1$ and the cosine similarity graph $\mathcal{G}_2$. 
In both graphs, $\mathcal{V}$ denotes the set of cell nodes, and $\mathcal{E}$ represents the edges indicating similarity between cells. 
Notably, the adjacency matrix of $\mathcal{G}_1$, $\mathbf{A}_1$, is computed via the inner product of $\mathbf{X}$, capturing feature-space similarity between cells, formulated as $\mathbf{A}_1=\mathbf{X}\cdot\mathbf{X}^T$. 
Concurrently, the adjacency matrix of $\mathcal{G}_2$, $\mathbf{A}_2$, is constructed based on cosine similarity, which is computed for each pair of cells $x_{i}$ and $x_{j}$ using 
$\cos \left(x_i, x_j\right)=\frac{x_i \cdot x_j}{\left\|x_i\right\|\left\|x_j\right\|}$, 
representing the angular similarity between the vectors of gene expression profiles of the two cells. 
Subsequently, additional matrix multiplication is applied to emphasize the similarity information and capture the directional differences in gene expression patterns between cells, defined as:
\begin{equation}
A_2=|\cos (\mathbf{X}, \mathbf{X})| \cdot|\cos (\mathbf{X}, \mathbf{X})|^T.
\label{equ:A_2}
\end{equation}
For graphs $\mathcal{G}_1$ and $\mathcal{G}_2$, $\mathbf{D}_1$, $\mathbf{D}_2$ are their diagonal matrices respectively. 
This dual-channel graph construction enables the effective capture of both feature-based and angular similarities between cells, improving the model's robustness to high-dimension, high-sparsity, and dropout events inherent in scRNA-seq data, thus facilitating more accurate cellular representation and downstream clustering tasks. 

To effectively integrate continuous cell similarities within the dual-channel soft graph, we employ a strategy that minimizes the Joint Normalized Cut (Joint NCut). 
This approach involves calculating the normalized laplacian matrices, $\mathbf{L}_1$ and $\mathbf{L}_2$, for the two soft graphs,
which together capture continuous structural information to identify optimal partitioning points in high-dimensional space: 
\begin{equation}
\begin{aligned}
      & \mathbf{L}_1 = \mathbf{I} - \mathbf{D}_1^{-1/2}\mathbf{A}_1\mathbf{D}_1^{-1/2} ,\\
      & \mathbf{L}_2 = \mathbf{I} - \mathbf{D}_2^{-1/2}\mathbf{A}_2\mathbf{D}_2^{-1/2}.
\label{equ:laplace}
\end{aligned}
\end{equation}

To fuse the similarity information from both soft graphs, we minimize the following Joint NCut objective:
\begin{equation}
\begin{aligned}
\text{Joint NCut}\left(\mathcal{V}\right) = \frac{1}{2} \sum_{i=1}^k \left( \frac{\text{cut}\left(\mathcal{V}_i, \mathcal{V} \setminus \mathcal{V}_i; \mathbf{L}_1\right)}{\text{vol}\left(\mathcal{V}_i; \mathbf{A}_1\right)} + \frac{\text{cut}(\mathcal{V}_i, \mathcal{V} \setminus \mathcal{V}_i; \mathbf{L}_2)}{\text{vol}\left(\mathcal{V}_i; \mathbf{A}_2\right)} \right), 
\label{equ:joint ncut}
\end{aligned}
\end{equation}
$\mathcal{V}_i$ refers to the set of nodes in cluster $i$. 
The term $\text{cut}\left(\mathcal{V}_i, \mathcal{V} \setminus \mathcal{V}_i\right)$ represents the weight of edges cut between the cluster $\mathcal{V}_i$ and the rest of the graph, while $\text{vol}\left(\mathcal{V}_i\right)$ is the volume of cluster $\mathcal{V}_i$. 
By jointly optimizing normalized cuts for both graphs, we achieve a consistent and optimized cellular representation. This strategy integrates continuous structural information from multiple similarity measures, enhancing the model's robustness and accuracy in handling high-dimensional, sparse scRNA-seq data.

Next, we formulate the objective function to minimize the Joint NCut, namely:
\begin{equation} 
\begin{aligned} & \arg\min_{\mathbf{Z}} \text{Tr}\left(\mathbf{Z}^{T}(\alpha \mathbf{L}_{1} + (1-\alpha)\mathbf{L}_{2})\mathbf{Z}\right) \ & \text{subject to } \mathbf{Z}^{T}\mathbf{Z} = \mathbf{I} , 
\label{equ:opt_ncut} 
\end{aligned} 
\end{equation}
where $\alpha$ is a weight parameter that balances the contributions of the two graphs to the optimization outcome. 
To ensure orthogonality in the cellular representations, we incorporate a regularization term, resulting in our loss function:
\begin{equation} 
\mathcal{L}_{\text{NCut}} = \text{Tr}\left(\mathbf{Z}^{T}(\alpha \mathbf{L}_{1} + (1-\alpha)\mathbf{L}_{2})\mathbf{Z}\right) + \beta |\mathbf{Z}^{T}\mathbf{Z} - \mathbf{I}|{F}.
\label{equ} 
\end{equation}

Here, $\beta$ is a regularization parameter and $\|\cdot\|_{F}$ denotes the Frobenius norm. 

Note that our method promotes the clustering of scRNA-seq data from multiple perspectives. 
First, the dual-channel soft graph construction effectively captures the continuous similarities between cells, enhancing robustness against high-dimension and high-sparsity. 
Second, the Joint NCut strategy integrates cellular relationships across similarity measures to improve clustering accuracy, 
while the optimized loss function ensures consistency of the learned embeddings across diverse graph structures, further enhancing performance.

\subsubsection{\textbf{Optimization via optimal transport}}
To ensure stable clustering results in complex data structures, we propose an optimal transport-based clustering optimization module that optimizes clustering assignments by minimizing the transportation cost. 
The objective is to minimize the cost of transferring the probability mass from an initial clustering assignment, represented by matrix $Q = [q_{ij}]$, to an optimal target distribution $P = [p_{ij}]$. 
Here, $q_{ij}$ denotes the probability of assigning cell $i$ to cluster $j$, and $P$ represents the optimal transport plan. 
The transport optimization problem is formulated as minimizing the divergence between $Q$ and $P$, constrained by the requirement that the cluster assignments respect the underlying data distribution. 
Specifically, we aim to minimize the cost matrix $-\log Q$, which quantifies the divergence between the observed data and the cluster centers while ensuring a balanced distribution of cluster assignments:
\begin{equation}
\vspace{-2mm}
\begin{aligned}
        \min_{P} \quad & -P*(\text{log}Q) \\
        \mbox{s.t.}\quad & P\in\mathbb{R}_{+}^{N\times C}, \\
         \quad &P\ \mathbf{1}_C = \mathbf{1}_N \ \text{ and } \ P^{T}\mathbf{1}_N = N\mathbf{\pi},
\end{aligned}
\label{equ:first_ot_optimization function}
\end{equation}
here, $\mathbf{1}_C$ and $\mathbf{1}_N$ are vectors of ones, and $\mathbf{\pi}$ represents the proportions of points in each cluster, which can be estimated from intermediate clustering results. 
The optimization ensures that the target distribution $P$ aligns with the proportions $\mathbf{\pi}$, preventing degenerate solutions where the clustering assignments become unbalanced. 
\begin{equation}
\vspace{-2mm}
\begin{aligned}
        \min_{P} \quad & -P*(\text{log}Q)-\frac{1}{\lambda}\text{H}(P) \\
        \mbox{s.t.}\quad & P\in\mathbb{R}_{+}^{N\times C}, \\
         \quad & P\ \mathbf{1}_C = \mathbf{1}_N \ \text{ and } \ P^{T}\mathbf{1}_N = N\mathbf{\pi},
\end{aligned}
\label{equ:ot_optimization function}
\end{equation}
where $H(P)$ is the entropy of the transport plan $P$, and $\lambda$ balances the accuracy of the assignment and the smoothness of the clustering. 
This optimization problem is solved through Sinkhorn’s fixed-point iteration method, which updates the transport matrix iteratively as follows: 
\begin{equation}
\vspace{-2mm}
\hat{P}^{(t)}=\operatorname{diag}\left(\boldsymbol{u}^{(t)}\right) Q^\lambda \operatorname{diag}\left(\boldsymbol{v}^{(t)}\right),
\label{equ:p(t)}
\end{equation}
where $\mathbf{u}^{(t)} = \mathbf{1}N / (Q^\lambda \mathbf{v}^{(t-1)})$ and $\mathbf{v} = N{\pi} / (Q^\lambda \mathbf{u}^{(t)})$. 
After several iterations, we obtain the optimal transport matrix $\hat{P}$.

Once $\hat{P}$ is fixed, we align $Q$ with $\hat{P}$, ensuring that the clustering assignment is consistent with the optimal transport plan. 
The clustering loss is defined as the Kullback-Leibler (KL) divergence between the estimated transport matrix $\hat{P}$ and the target distribution $Q$:
\begin{equation}
    \mathcal{L}_{\text{K L}}=\mathbf{\text{KL}}(\hat{P} \| Q).
\label{equ:loss_kl}
\end{equation}

Overall, this module improves clustering accuracy by balancing the assignments across clusters and maintaining consistency with the data structure. 
It enhances~\methodname's ability to handle high-dimensional, high-sparse scRNA-seq data, ensuring more stable and reliable clustering results.

\subsubsection{\textbf{Joint optimiztion}}
The overall optimization objective of~\methodname~combines three main components: NCut loss, ZINB loss, and clustering loss, formulated as:
\begin{equation}
\mathcal{L} = \mathcal{L}_{N C u t} + \gamma\mathcal{L}_{Z I N B } + \mu\mathcal{L}_{K L},
\label{equ:total_loss}
\end{equation}
where the coefficients $\gamma$ and $\mu$ serve as balancing factors. 
Each term plays a crucial role in fine-tuning~\methodname's performance.  

\subsection{\textbf{Datasets}}
We evaluated the performance of~\methodname~on 10 scRNA-seq datasets comprising both human and mouse samples, covering diverse cell types such as pancreas, liver, and muscle cells.
These datasets were generated using various sequencing platforms including Illumina HiSeq 2500, NextSeq 500, and NovaSeq 6000, and employed sequencing methods such as CEL-seq, 10X Genomics, and Smart-seq2.
The number of cells per dataset ranges from 822 to 8617, with gene counts varying between 2,000 and 61,759. The datasets include between 6 and 15 annotated cell groups, and exhibit sparsity rates ranging from 73.02\% to 95.42\%.
Detailed information for each dataset is summarized in Table\ref{tab:dataset}.

\begin{table*}[t!]
    \centering
    \caption{Summary of the scRNA-seq datasets.}
    \renewcommand\arraystretch{1.2}
    \setlength{\tabcolsep}{6mm}
    \resizebox{\textwidth}{!}{
        \begin{tabular}{lccccccc}
        \toprule
        \textbf{Datasets} & \textbf{Sequencing platform} & \makecell{\textbf{Sequencing} \\ \textbf{method}} & \makecell{\textbf{Sample} \\ \textbf{size}} & \makecell{\textbf{No. of} \\ \textbf{genes}} & \makecell{\textbf{No. of} \\ \textbf{groups}} & \textbf{Cluster sizes} & \makecell{\textbf{Sparsity} \\ \textbf{rate(\%)}} \\
        \midrule
        Mouse Pancreas cells 1~\cite{baron2016single} & Illumina HiSeq 2500 & CEL-seq & 822 & 14878 & 13 & (343,236,85,3,29,72,14,9,17,4,4,2,4) & 90.48 \\
        Mouse Pancreas cells 2~\cite{baron2016single} & Illumina HiSeq 2500 & CEL-seq & 1064 & 14878 & 13 & (551,39,133,182,27,67,8,19,4,10,18,3,3) & 87.81 \\
        Human Pancreas cells 2~\cite{baron2016single} & Illumina HiSeq 2500 & CEL-seq & 1724 & 20125 & 14 & (125,676,81,301,371,17,22,86,23,2,6,9,3,2) & 90.59 \\
        Human Pancreas cells 1~\cite{baron2016single} & Illumina HiSeq 2500 & CEL-seq & 1937 & 20125 & 14 & (110,872,214,51,120,236,13,70,130,92,14,5,8,2) & 90.45 \\
        Muraro Human Pancreas cells~\cite{muraro2016single} & Illumina NextSeq 500 & CEL-seq2 & 2122 & 19046 & 9 & (219,812,448,193,245,21,3,101,80) & 73.02 \\
        Human Pancreas cells 3~\cite{baron2016single} & Illumina HiSeq 2500 & CEL-seq & 3605 & 20125 & 14 & (843,161,376,130,787,92,100,54,36,14,7,2,2,1) & 91.30 \\
        CITE-CMBC~\cite{zheng2017massively} & Illumina NextSeq 500 & 10X Genomics & 8617 & 2000 & 15 & (1791,2293,1248,1089,608,350,273,230,182,119,114,105,96,70,49) & 93.26 \\
        Human Liver cells~\cite{macparland2018single} & Illumina HiSeq 2500 & 10X Genomics & 8444 & 4999 & 11 & (844,119,1192,961,488,569,3501,129,37,511,93) & 90.77 \\
        Tabula Muris Limb Muscle cells~\cite{tabula2020single} & Illumina NovaSeq 6000 & Smart-seq2 & 3855 & 21069 & 6 & (1833,935,453,311,187,136) & 91.38 \\
        Tabula Sapiens Liver cells~\cite{tabula2020single} & -- & Smart-seq2 & 2152 & 61759 & 15 & (741,323,177,172,129,98,81,77,74,74,66,56,33,29,22) & 95.42 \\
        \bottomrule
        \end{tabular}
    }
    \label{tab:dataset}
\end{table*}


\subsection{\textbf{Baselines}}
To highlight the advantages of our method, we compared it with 11 prominent existing approaches, including traditional clustering methods, deep learning-based clustering algorithms, and deep structured clustering approaches. Among the \textbf{traditional methods}, \textbf{pcaReduce}~\cite{vzurauskiene2016pcareduce} performs clustering by iteratively combining probability density functions, and \textbf{SUSSC}~\cite{wang2021suscc} uses stochastic gradient descent and nearest neighbor search for clustering.

In the realm of \textbf{deep learning clustering}, we considered several foundational models. 
\textbf{DEC} utilizes deep embedded clustering, employing neural networks to simultaneously learn feature representations and cluster assignments~\cite{xie2016unsupervised}. 
contrastive-sc adapts self-supervised contrastive learning to scRNA-seq data, decoupling embedding learning and clustering for flexible and robust cell type identification~\cite{ciortan2021contrastive}. The method using contrastive-sc embeddings followed by K-Means clustering is referred to as \textbf{contrastive-sc K-means}, while that using Leiden clustering is referred to as \textbf{contrastive-sc Leiden}.
\textbf{scDeepCluster} leverages the ZINB model to simulate the distribution of scRNA-seq data while concurrently learning feature representations and clustering~\cite{tian2019clustering}. 
\textbf{scDSSC} enhances clustering performance through explicit modeling of scRNA-seq data generation~\cite{wang2022scdssc}. 
Additionally, \textbf{scNAME} integrates mask estimation and neighborhood contrastive learning to achieve denoising and clustering~\cite{wan2022scname}, 
while \textbf{AttentionAE-sc} employs attention mechanisms to enhance representation learning~\cite{li2023attention}. 
\textbf{scziDesk} introduces a denoising autoencoder and optimizes the clustering results in the latent space through a soft self-training K-means algorithm~\cite{chen2020deep}.

For \textbf{deep structured clustering methods}, we compared scGNN~\cite{wang2021scgnn}, scDSC~\cite{gan2022deep}, and scCDCG~\cite{xu2024sccdcg}. 
\textbf{scGNN} utilizes GNNs and multi-modal autoencoders to capture intercellular relationships and processes heterogeneous gene expression patterns through a left-truncated mixture Gaussian model~\cite{wang2021scgnn}. 
\textbf{scDSC} combines a ZINB model with GNN modules for data representation learning and uses a mutual supervision strategy to unify these architectures~\cite{gan2022deep}. 
Lastly, \textbf{scCDCG} effectively handles high-dimensional and high-sparsity scRNA-seq data by leveraging higher-order structural information, demonstrating exceptional efficiency~\cite{xu2024sccdcg}.

\section{\textbf{Results}}
\subsection{\textbf{Experimental Setup}}
\subsubsection{\textbf{Dataset Preprocessing}}
Raw scRNA-seq data were preprocessed using the SCANPY package~\cite{wolf2018scanpy}, resulting in the processed dataset $\mathbf{X}$ as input for~\methodname. 
Initially, genes expressed in fewer than three cells were filtered out, and genes with zero counts across all cells were removed. 
The remaining data were then normalized and log-transformed using TPM. 
Finally, the top $n$ most highly expressed genes were selected from the remaining genes to construct the preprocessed dataset, with the specific value of $n$ for each dataset outlined in Table~\ref{tab:Hyperparameter_settings}.

For baseline methods, we adhered to the preprocessing steps specified in their respective publications to ensure consistency with recommended practices and allow for fair comparison while preserving the integrity of each method's design.

\subsubsection{\textbf{Evaluation Metrics}}
To assess the efficacy of the proposed method, we evaluate its clustering performance using three widely recognized metrics form public domain: Accuracy (ACC), Normalized Mutual Information (NMI)~\cite{strehl2002cluster}, and Adjusted Rand Index (ARI)~\cite{vinh2009information}. 
Higher values for these metrics indicate better clustering results. 

\subsubsection{\textbf{Training Procedure}}
The training process of~\methodname~includes two steps:
(i) we pre-train~\methodname~with both NCut loss and ZINB loss by 200 epochs;
(ii) under the guidance of Eq.~\ref{equ:total_loss}, we train the whole network with 200 epochs. 
As the number of cell clusters $k$ is unknown in real applications, we apply K-means to the embeddings to obtain the optimal value of $k$.


\begin{table}[t!]
\caption{Hyperparameter settings for different datasets. The symbol `--' indicates that this operation was not performed for the respective dataset.}
    \centering
    \begin{tabular}{lccccc}
         \toprule
         Datasets & $\alpha$  & $\beta$ & $\gamma$ & \makecell{Weight\\ decay} & \makecell{No. of\\highly\\expression\\ genes}\\
         \midrule
         Mouse Pancreas cells 2 & 0.5 & 20 & 25 & 1e-3 & 1500 \\
         Mouse Pancreas cells 1 & 0.7 & 25 & 20 & 5e-3 & 1500 \\
         Human Pancreas cells 2 & 0.7 & 10 & 25 & 5e-4 & 2000 \\
         Human Pancreas cells 1 & 0.7 & 30 & 25 & 5e-4 & 2000 \\
         Mauro Human Pancreas cells & 0.5 & 10 & 100 & 5e-3 & 2000 \\
         Human Pancreas cells 3 & 0.1 & 30 & 25 & 5e-3 & 2000 \\
         CITE-CMBC & 0.7 & 10 & 50 & 5e-3 & 1500 \\
         Human Liver cells & 0.7 & 15 & 25 & 5e-3 & 1500 \\
         Tabula  Muris Limb Muscle cells & 0.75 & 20 & 25 & 1e-4 & -- \\ 
         Tabula Spaies Liver cells & 0.5 & 20 & 100 & 5e-3 & 8000 \\
         \bottomrule
    \end{tabular}
    \label{tab:Hyperparameter_settings}
\end{table}

\subsubsection{\textbf{Parameters Setting}}
In practice, we implement~\methodname~in Python 3.7 based on PyTorch.  
The embedding size $d$ is fixed at 16 for all datasets.
Furthermore, we conduct meticulous parameter tuning across different datasets to ensure optimal model performance. 
The specific parameter settings are as follows: 
the model training is optimized using the Adam optimizer with a uniformly set learning rate of 1e-3 across all datasets and supplemented by a weight decay strategy to further enhance~\methodname's generalization capability, ensuring stable convergence; for all datasets, the hyperparameters $\lambda$, $\theta$ and $\mu$ are initially set to 5, 1, and 1e-3, respectively; 
for additional details on the other hyperparameter settings, please see Table~\ref{tab:Hyperparameter_settings}. 

For baseline methods, we strictly followed the parameter settings provided in their respective publications. In cases where parameter settings were not clearly specified, we conducted preliminary experiments to determine the optimal values, ensuring each tool operated under its best conditions.  

To ensure the accuracy of the experimental results, we repeat each experiment 10 times with varying random seeds across all datasets, reporting the mean and variance of the outcomes.

\subsection{\textbf{Overall Performance}}
\subsubsection{\textbf{Quantitative Analysis}}

Table~\ref{tab:expriment} presents a comparative analysis of clustering performance across ten benchmark datasets, evaluating different models based on ACC, NMI, and ARI. 
The results highlight three key observations:
(i)~\methodname~outperforms all other models across all three metrics, demonstrating its effectiveness in tackling scRNA-seq clustering challenges. 
On average,~\methodname~exhibits improvements of 4.06\%, 3.725\%, and 4.03\% in ACC, NMI, and ARI, respectively, when compared to the second-best approach, highlighting its comprehensive advantage across multiple evaluation dimensions. 
(ii) Deep learning-based methods, such as Attention-AE, primarily rely on node features while disregarding complex structural relationships, making them ineffective in handling the high sparsity and dropout effects in scRNA-seq data.
\methodname~addresses this by integrating a ZINB-based feature autoencoder to enhance feature extraction and an optimal transport-based clustering module to refine cluster assignments by minimizing transport cost, leading to more stable and biologically meaningful clustering.
(iii) GNN-based methods, which rely on predefined hard graph structures, often struggle with representation collapse and fail to capture the continuous similarities between cells. This limitation arises because hard graphs enforce binary connections, leading to excessive discretization and the loss of nuanced intercellular relationships. 
In contrast,~\methodname~introduces a dual-channel soft graph construction with a minimized Joint NCut strategy, effectively capturing continuous cell relationships, reducing rigid graph dependencies, and improving representation learning.


\begin{table*}[t!]
    \centering
    \caption{Clustering performances of all models on 10 benchmark datasets (mean $\pm$ standard). The \textbf{bold} values represent the best results, and \underline{underline} values represent the runner-up results.}
    \resizebox{\textwidth}{!}{ 
    \renewcommand\arraystretch{1.5}
    \begin{tabular}{p{3.4cm}|p{1.3cm}|p{2.3cm}p{2cm}p{2cm}p{2.3cm}p{2.3cm}p{2cm}p{2cm}p{2cm}p{2cm}p{2cm}|p{2cm}p{2cm}p{2cm}|p{2.5cm}}
    
        \toprule
        \textbf{\large Datasets} & \textbf{\large Metric} & \textbf{\large \makecell{\large pcaReduce}} & \textbf{\large \makecell{SUSSC}} & \textbf{\large \makecell{DEC}} & \textbf{\large \makecell{contrastive\\-sc\\K-means}} & \textbf{\large \makecell{contrastive\\-sc\\Leiden}} & \textbf{\large \makecell{scDeep-\\Cluster}} & \textbf{\large scDSSC} & \textbf{\large scNAME} & \textbf{\large scziDesk} & \textbf{\large \makecell{Attention\\-AE}} & \textbf{\large scDSC} & \textbf{\large scGNN} & \textbf{\large \makecell{scCDCG}}& \textbf{\large \makecell{\methodname}}\\
        \midrule
        \multirow{3}*{\textbf{\large \makecell{Mouse Pancreas\\ cells 1}}}
        & \large ACC & \large 42.50 \small$\pm$ 0.1 & \large 40.00 \small$\pm$ 0.8 &\large 47.79 \small$\pm$ 0.7 & \large 57.58 \small$\pm$ 4.2 & \large 57.53 \small$\pm$ 4.7 & \large 65.18 \small$\pm$ 0.7 &\large 71.51 \small$\pm$ 11.3 & \large 75.05 \small$\pm$ 5.5 &\large 78.73 \small$\pm$ 4.3 & \large 81.70 \small$\pm$ 4.2 & \large 69.91 \small$\pm$ 3.8 & \large 47.93 \small$\pm$ 4.9 & \large \underline{82.64} \small$\pm$ 0.5 & \large \textbf{91.24} \small$\pm$ 1.0\\
        & \large NMI &  \large 47.19 \small$\pm$ 0.1 & \large 61.61 \small$\pm$ 0.1 & \large55.31 \small$\pm$ 1.9 & \large 65.62 \small$\pm$ 2.1 & \large 57.20 \small$\pm$ 14.0 &\large 65.92 \small$\pm$ 0.9 &\large 75.63 \small$\pm$ 4.6 &\large 72.25 \small$\pm$ 1.2 &\large 73.00 \small$\pm$ 7.5 & \large \underline{79.92} \small$\pm$ 3.0 &\large 62.04 \small$\pm$ 5.4 & \large 58.77 \small$\pm$ 1.7 & \large 77.66 \small$\pm$ 2.0 & \large \textbf{85.78} \small$\pm$ 0.5\\
        & \large ARI &  \large 18.32 \small$\pm$ 0.1 & \large 30.73 \small$\pm$ 1.0 & \large33.04 \small$\pm$ 2.3 & \large 47.01 \small$\pm$ 4.3 & \large 55.96 \small$\pm$ 9.3 & \large 55.47 \small$\pm$ 2.2 &\large 62.88 \small$\pm$ 11.3 &\large 64.28 \small$\pm$ 6.9 &\large 75.24 \small$\pm$ 7.5 & \large 69.94 \small$\pm$ 5.05 &\large 64.78 \small$\pm$ 8.4 & \large 36.33 \small$\pm$ 5.5 & \large \underline{80.60} \small$\pm$ 5.8 & \large \textbf{91.63} \small$\pm$ 1.0\\
        \hline

        \multirow{3}*{\textbf{\large \makecell{Mouse Pancreas \\ cells 2}}}
        & \large ACC &  \large 28.97 \small$\pm$ 0.3 & \large 36.41 \small$\pm$ 0.1 &\large 34.19 \small$\pm$ 2.3 & \large 59.55 \small$\pm$ 5.7 & \large 47.61 \small$\pm$ 3.6 & \large 69.91 \small$\pm$ 2.9 &\large 89.11 \small$\pm$ 1.4 & \large 82.36 \small$\pm$ 8.0 &\large 85.71 \small$\pm$ 2.0 & \large 81.92 \small$\pm$ 6.5 & \large 77.35 \small$\pm$ 2.3 & \large 43.03 \small$\pm$ 3.0 & \large \underline{93.97} \small$\pm$ 0.2 & \large \textbf{94.45} \small$\pm$ 0.2\\
        & \large NMI &  \large 48.93 \small$\pm$ 0.1 & \large 60.31 \small$\pm$ 0.2 & \large54.08 \small$\pm$ 0.9 & \large 62.34 \small$\pm$ 2.4 & \large 53.42 \small$\pm$ 11.0 &\large 59.58 \small$\pm$ 2.6 &\large 83.75 \small$\pm$ 1.7 &\large 78.00 \small$\pm$ 3.9 &\large 79.91 \small$\pm$ 2.1 & \large 81.88 \small$\pm$ 5.1 &\large 62.50 \small$\pm$ 0.4 & \large 55.87 \small$\pm$ 1.9 & \large \underline{88.02} \small$\pm$ 0.2 & \large \textbf{88.45} \small$\pm$ 0.9\\
        & \large ARI &  \large 12.80 \small$\pm$ 0.1 & \large 24.90 \small$\pm$ 0.1 & \large18.54 \small$\pm$ 1.1 & \large 49.75 \small$\pm$ 8.5 & \large 37.88 \small$\pm$ 13.1 & \large 60.14 \small$\pm$ 3.6 &\large 89.66 \small$\pm$ 0.8 &\large 76.56 \small$\pm$ 11.4 &\large 81.19 \small$\pm$ 4.3 & \large 82.14 \small$\pm$ 7.8 & \large 65.81 \small$\pm$ 1.4 & \large 26.40 \small$\pm$ 2.5 & \large \underline{92.61} \small$\pm$ 0.3 & \large \textbf{92.64} \small$\pm$ 1.0\\
        \hline

        \multirow{3}*{\textbf{\large \makecell{Human Pancreas\\ cells 2}}}
        & \large ACC & \large 37.56 \small$\pm$ 0.4 & \large 46.75 \small$\pm$ 0.2 & \large 45.26 \small$\pm$ 1.8 & \large 60.01 \small$\pm$ 5.7 & \large 50.31 \small$\pm$ 2.0 & \large 62.88 \small$\pm$ 5.5 & \large 80.02 \small$\pm$ 9.4 & \large 63.33 \small$\pm$ 2.0 & \large 72.72 \small$\pm$ 6.2 & \large 81.80 \small$\pm$ 12.1 & \large 80.40 \small$\pm$ 4.0 & \large 57.62 \small$\pm$ 2.3 & \large \underline{84.66} \small$\pm$ 1.1 & \large \textbf{91.77} \small$\pm$ 0.5\\
        & \large NMI & \large 50.81 \small$\pm$ 0.3 & \large 68.62 \small$\pm$ 0.1 & \large 66.32 \small$\pm$ 2.1 & \large 62.25 \small$\pm$ 2.3 & \large 51.99 \small$\pm$ 11.2 & \large 73.97 \small$\pm$ 2.2 & \large 75.34 \small$\pm$ 5.5 & \large 58.05 \small$\pm$ 1.6 & \large 76.42 \small$\pm$ 3.9 & \large 82.16 \small$\pm$ 10.9 & \large 79.44 \small$\pm$ 1.7 & \large 79.32 \small$\pm$ 2.3 & \large \underline{83.44} \small$\pm$ 2.6 & \large \textbf{91.21} \small$\pm$ 0.5\\
        & \large ARI & \large 19.23 \small$\pm$ 0.5 & \large 38.45 \small$\pm$ 0.4 & \large39.14 \small$\pm$ 1.8 & \large 47.10 \small$\pm$ 5.3 & \large 47.12 \small$\pm$ 13.2 & \large 64.56 \small$\pm$ 7.5 & \large 67.06 \small$\pm$ 14.0 & \large 32.12 \small$\pm$ 1.5 &\large 57.70 \small$\pm$ 9.7 & \large 73.20 \small$\pm$ 10.5 & \large 82.85 \small$\pm$ 6.5 & \large 79.96 \small$\pm$ 1.7 & \large \underline{85.30} \small$\pm$ 1.6 & \large \textbf{92.62} \small$\pm$ 0.4\\
        \hline

        \multirow{3}*{\textbf{\large \makecell{Human Pancreas \\cells 1}}}
        & \large ACC & \large 23.63 \small$\pm$ 0.2 & \large 40.65 \small$\pm$ 0.1 & \large 40.64 \small$\pm$ 1.9 & \large 69.67 \small$\pm$ 7.8 & \large 60.68 \small$\pm$ 2.8 & \large 66.39 \small$\pm$ 3.8 & \large 79.29 \small$\pm$ 9.6 & \large 84.20 \small$\pm$ 6.0 & \large 86.20 \small$\pm$ 7.0 & \large 81.56 \small$\pm$ 1.8 & \large 73.10 \small$\pm$ 2.3 & \large 55.10 \small$\pm$ 2.1 & \large \underline{92.15} \small$\pm$ 0.8 & \large \textbf{96.25} \small$\pm$ 0.8\\
        & \large NMI & \large 42.58 \small$\pm$ 0.4 & \large 67.95 \small$\pm$ 0.1 & \large62.35 \small$\pm$ 0.6 & \large 69.21 \small$\pm$ 3.3 & \large 65.61 \small$\pm$ 2.3 & \large 69.42 \small$\pm$ 12.5 & \large 83.84 \small$\pm$ 6.4 & \large 65.74 \small$\pm$ 4.0 & \large 84.99 \small$\pm$ 2.5 & \large 82.62 \small$\pm$ 3.0 & \large 60.50 \small$\pm$ 4.2 & \large 64.03 \small$\pm$ 0.9 & \large \underline{86.35} \small$\pm$ 0.9 & \large \textbf{91.42} \small$\pm$ 0.9\\
        & \large ARI & \large 1.98 \small$\pm$ 0.5 & \large 31.16 \small$\pm$ 0.1 & \large 26.07 \small$\pm$ 1.3 & \large 52.00 \small$\pm$ 8.7 & \large 37.11 \small$\pm$ 4.0 & \large 47.92 \small$\pm$ 3.0 & \large 72.07 \small$\pm$ 14.3 & \large 63.31 \small$\pm$ 7.4 & \large 78.16 \small$\pm$ 11.6 & \large 71.84 \small$\pm$ 8.5 & \large 69.81 \small$\pm$ 1.6 & \large 38.32 \small$\pm$ 0.8 & \large \underline{92.83} \small$\pm$ 0.6 & \large \textbf{94.89} \small$\pm$ 0.2\\
        \hline
        
        \multirow{3}*{\textbf{\large \makecell{Mauro Human\\ Pancreas cells}}}
        & \large ACC & \large 50.56 \small$\pm$ 3.4 & \large 76.26 \small$\pm$ 0.1 & \large 72.22 \small$\pm$ 5.5 & \large 81.85 \small$\pm$ 4.6 & \large 67.22 \small$\pm$ 4.2 & \large 74.70 \small$\pm$ 2.7 & \large 80.10 \small$\pm$ 4.8 & \large 80.36 \small$\pm$ 7.9 & \large 91.51 \small$\pm$ 8.9 & \large \underline{93.84} \small$\pm$ 2.7 & \large 79.42 \small$\pm$ 1.4 & \large 79.32 \small$\pm$ 4.0 & \large 92.65 \small$\pm$ 1.9 & \large \textbf{96.03} \small$\pm$ 0.2\\
        & \large NMI & \large 54.83 \small$\pm$ 1.7 & \large 82.25 \small$\pm$ 0.00 & \large76.29 \small$\pm$ 2.7 & \large 77.73 \small$\pm$ 3.3 & \large 75.31 \small$\pm$ 1.7 & \large 79.32 \small$\pm$ 0.3 & \large 82.16 \small$\pm$ 4.7 & \large 79.12 \small$\pm$ 2.7 & \large 86.49 \small$\pm$ 4.6 & \large \underline{87.80} \small$\pm$ 2.3 & \large 75.89 \small$\pm$ 0.6 & \large 78.76 \small$\pm$ 5.6 & \large 86.81 \small$\pm$ 1.0 & \large \textbf{90.23} \small$\pm$ 1.7\\
        & \large ARI &  \large 36.61 \small$\pm$ 2.3 & \large 64.23 \small$\pm$ 0.1 & \large61.76 \small$\pm$ 5.6 & \large 75.22 \small$\pm$ 10.1 & \large 60.04 \small$\pm$ 4.9 & \large 64.59 \small$\pm$ 2.5 & \large 85.76 \small$\pm$ 7.3 & \large 76.64 \small$\pm$ 10.4 & \large 88.03 \small$\pm$ 11.3 & \large 90.74 \small$\pm$ 4.1 & \large 75.42 \small$\pm$ 1.2 & \large 78.87 \small$\pm$ 3.2 & \large \underline{91.37} \small$\pm$ 1.2 & \large \textbf{92.39} \small$\pm$ 1.9\\
        \hline
        
        \multirow{3}*{\textbf{\large \makecell{Human Pancreas\\ cells 3}}}
        & \large ACC & \large 35.92 \small$\pm$ 0.4 & \large 45.16 \small$\pm$ 0.1 & \large 43.56 \small$\pm$ 1.5 & \large 57.47 \small$\pm$ 5.0 & \large 52.77 \small$\pm$ 3.6 & \large73.65 \small$\pm$ 3.1 & \large 71.61 \small$\pm$ 9.3 & \large 83.20 \small$\pm$ 8.7 & \large 84.13 \small$\pm$ 9.3  & \large \underline{90.96} \small$\pm$ 2.4 & \large 80.12 \small$\pm$ 10.0 & \large 67.94 \small$\pm$ 4.6 & \large 88.04 \small$\pm$ 0.3 & \large \textbf{94.80} \small$\pm$ 0.8\\
        & \large NMI & \large 51.22 \small$\pm$ 0.3 & \large 69.50 \small$\pm$ 0.1 & \large64.51 \small$\pm$ 0.8 & \large 66.84 \small$\pm$ 2.3 & \large 67.39 \small$\pm$ 2.5 & \large 75.02 \small$\pm$ 1.2 & \large 75.77 \small$\pm$ 6.2 & \large 80.56 \small$\pm$ 5.6 & \large 82.02 \small$\pm$ 3.6 & \large \underline{87.00} \small$\pm$ 1.8 & \large 75.87 \small$\pm$ 8.4 & \large 62.60 \small$\pm$ 1.8 & \large 82.21 \small$\pm$ 1.1 & \large \textbf{88.81} \small$\pm$ 0.5\\
        & \large ARI & \large 26.40 \small$\pm$ 4.5 & \large 43.53 \small$\pm$ 0.1 & \large42.73 \small$\pm$ 1.0 & \large 47.9 \small$\pm$ 3.4 & \large 46.18 \small$\pm$ 3.9 & \large 63.31 \small$\pm$ 2.3 & \large 64.86 \small$\pm$ 13.0 & \large 81.08 \small$\pm$ 5.7 & \large 72.77 \small$\pm$ 10.9 & \large 89.78 \small$\pm$ 2.4 & \large 82.85 \small$\pm$ 8.5 & \large 58.41 \small$\pm$ 1.9 & \large \underline{90.78} \small$\pm$ 0.6 & \large \textbf{93.58} \small$\pm$ 0.1\\
        \hline

        \multirow{3}*{\textbf{\large \makecell{CITE-CMBC}}}
        & \large ACC & \large 28.19 \small$\pm$ 0.1 & \large 47.22 \small$\pm$ 0.1 & \large 41.88 \small$\pm$ 5.4 & \large 52.70 \small$\pm$ 5.0 & \large 49.84 \small$\pm$ 3.7 & \large 70.80 \small$\pm$ 2.5 & \large 63.44 \small$\pm$ 8.5 &\large 68.06 \small$\pm$ 13.3 & \large 70.70 \small$\pm$ 2.7 & \large 61.18 \small$\pm$ 7.9 & \large 68.79 \small$\pm$ 0.87 & \large 66.71 \small$\pm$ 4.0 & \large \underline{71.45} \small$\pm$ 1.8 & \large \textbf{75.89} \small$\pm$ 0.3\\
        & \large NMI & \large 28.36 \small$\pm$ 0.2 & \large 60.14 \small$\pm$ 0.1 & \large46.87 \small$\pm$ 9.2 & \large 64.83 \small$\pm$ 0.8 & \large 63.57 \small$\pm$ 8.7 & \large 72.53 \small$\pm$ 0.7 & \large 64.04 \small$\pm$ 9.9 & \large 63.23 \small$\pm$ 13.2 & \large 66.97 \small$\pm$ 1.9 & \large 62.02 \small$\pm$ 9.6 & \large 64.21 \small$\pm$ 3.3 & \large 61.70 \small$\pm$ 3.1 & \large \underline{74.77} \small$\pm$ 1.7 & \large \textbf{76.18} \small$\pm$ 3.6\\
        & \large ARI & \large 5.10 \small$\pm$ 0.1 & \large 35.56 \small$\pm$ 1.6 & \large 23.55 \small$\pm$ 10.3 & \large 47.88 \small$\pm$ 3.7 & \large 52.27 \small$\pm$ 8.8 & \large 56.23 \small$\pm$ 4.1 & \large 55.28 \small$\pm$ 10.9 & \large 58.99 \small$\pm$ 21.3 & \large 56.38 \small$\pm$ 3.8 & \large 44.20 \small$\pm$ 11.6 & \large 52.51 \small$\pm$ 2.2 & \large 60.25 \small$\pm$ 1.8 & \large \underline{61.46} \small$\pm$ 1.4 & \large \textbf{64.47} \small$\pm$ 0.9\\
        \hline

        \multirow{3}*{\textbf{\large \makecell{Human Liver\\ cells}}}
        & \large ACC &  \large 34.92 \small$\pm$ 0.2 & \large 46.62 \small$\pm$ 1.6 & \large 46.32 \small$\pm$ 5.5 & \large 79.31 \small$\pm$ 2.1 & \large 47.18 \small$\pm$ 6.4 & \large 63.44 \small$\pm$ 0.7 & \large 70.60 \small$\pm$ 7.1 & \large 74.55 \small$\pm$ 5.3 & \large 74.01 \small$\pm$ 8.3 & \large \underline{79.46} \small$\pm$ 6.7 & \large 70.90 \small$\pm$ 2.2 & \large 68.65 \small$\pm$ 3.2 & \large 75.34 \small$\pm$ 1.7 & \large \textbf{91.16} \small$\pm$ 0.3\\
        & \large NMI &  \large 32.07 \small$\pm$ 0.6 & \large 67.32 \small$\pm$ 0.0 & \large52.24 \small$\pm$ 5.0 & \large \underline{85.11} \small$\pm$ 0.1 & \large 75.48 \small$\pm$ 1.4 & \large 74.60 \small$\pm$ 2.5 & \large 78.94 \small$\pm$ 11.3 & \large 77.42 \small$\pm$ 12.0 & \large 72.59 \small$\pm$ 3.7 & \large 82.86 \small$\pm$ 4.5 & \large 71.63 \small$\pm$ 2.8 & \large 62.33 \small$\pm$ 2.1 & \large 79.34 \small$\pm$ 2.6 & \large \textbf{89.17} \small$\pm$ 0.6\\
        & \large ARI &  \large 6.34 \small$\pm$ 14.4 & \large 35.12 \small$\pm$ 0.0 & \large31.04 \small$\pm$ 4.7 & \large \underline{89.96} \small$\pm$ 0.1 & \large 35.18 \small$\pm$ 4.1 & \large 58.58 \small$\pm$ 10.1 & \large 73.41 \small$\pm$ 11.8 & \large 79.91 \small$\pm$ 2.9 & \large 77.44 \small$\pm$ 12.9 & \large 79.16 \small$\pm$ 11.8 & \large 75.34 \small$\pm$ 3.6 & \large 65.41 \small$\pm$ 1.3 & \large 81.26 \small$\pm$ 2.7 & \large \textbf{93.65} \small$\pm$ 0.5\\
        \hline
        
        \multirow{3}*{\textbf{\large \makecell{Tabula Muris  \\ Limb Muscle cells}}}
        & \large ACC & \large 29.19 \small$\pm$ 1.1 & \large 29.71 \small$\pm$ 2.4 & \large 54.79 \small$\pm$ 7.2 & \large 46.49 \small$\pm$ 1.3 & \large 32.57 \small$\pm$ 4.5 & \large 59.57 \small$\pm$ 4.3 & \large 59.47 \small$\pm$ 4.7 &\large 61.34 \small$\pm$ 3.4 & \large 53.31 \small$\pm$ 4.9 & \large 53.35 \small$\pm$ 11.7 & \large 48.62 \small$\pm$ 2.5 & \large \underline{66.71} \small$\pm$ 4.0 & \large 62.25 \small$\pm$ 7.9 & \large \textbf{68.68} \small$\pm$ 2.3\\
        & \large NMI & \large 19.26 \small$\pm$ 1.2 & \large 8.6 \small$\pm$ 0.9 & \large 51.49 \small$\pm$ 8.0 & \large 31.71 \small$\pm$ 1.5 & \large 33.12 \small$\pm$ 0.9 & \large 34.55 \small$\pm$ 5.5 & \large 40.88 \small$\pm$ 0.34 & \large \underline{63.51} \small$\pm$ 1.4 & \large 36.66 \small$\pm$ 7.1 & \large 12.55 \small$\pm$ 6.8 & \large 22.11 \small$\pm$ 5.2 & \large 61.70 \small$\pm$ 3.1 & \large 56.54 \small$\pm$ 8.5 & \large \textbf{71.23} \small$\pm$ 0.7\\
        & \large ARI & \large 15.20 \small$\pm$ 1.1 & \large 5.86 \small$\pm$ 1.9 & \large 35.94 \small$\pm$ 11.2 & \large 21.26 \small$\pm$ 2.1 & \large 18.75 \small$\pm$ 2.1 & \large 35.93 \small$\pm$ 9.1 & \large 42.07 \small$\pm$ 5.7 & \large 54.70 \small$\pm$ 2.6 & \large 32.59 \small$\pm$ 8.1 & \large 13.64 \small$\pm$ 7.5 & \large 21.90 \small$\pm$ 3.6 & \large \underline{60.25} \small$\pm$ 1.8 & \large 53.37 \small$\pm$ 9.4 & \large \textbf{69.01} \small$\pm$ 1.9\\
        \hline

        \multirow{3}*{\textbf{\large \makecell{Tabula Spaies \\ Liver cells}}}
        & \large ACC &  \large 31.23 \small$\pm$ 5.3 & \large 41.23 \small$\pm$ 6.5 & \large 65.76 \small$\pm$ 3.6 & \large 45.50 \small$\pm$ 1.3 & \large 46.94 \small$\pm$ 3.5 & \large 49.08 \small$\pm$ 2.1 & \large 47.49 \small$\pm$ 4.7 & \large 63.33 \small$\pm$ 1.9 & \large 66.68 \small$\pm$ 3.3 & \large 53.28 \small$\pm$ 12.5 & \large \underline{73.83} \small$\pm$ 3.4 & \large 68.65 \small$\pm$ 3.2 & \large 48.86 \small$\pm$ 1.6 & \large \textbf{73.90} \small$\pm$ 0.6\\
        & \large NMI &  \large 41.34 \small$\pm$ 4.7 & \large 57.63 \small$\pm$ 5.3 & \large 71.92 \small$\pm$ 2.1 & \large 48.48 \small$\pm$ 12.8 & \large 54.89 \small$\pm$ 2.2 & \large 60.98 \small$\pm$ 2.5 & \large 44.56 \small$\pm$ 3.4 & \large 74.83 \small$\pm$ 0.8 & \large 75.10 \small$\pm$ 1.3 & \large 51.17 \small$\pm$ 9.4 & \large \underline{77.21} \small$\pm$ 2.2 & \large 62.33 \small$\pm$ 2.1 & \large 62.54 \small$\pm$ 3.6 & \large \textbf{77.83} \small$\pm$ 2.0\\
        & \large ARI &  \large 33.40 \small$\pm$ 6.7 & \large 28.82 \small$\pm$ 5.4 & \large 65.47 \small$\pm$ 4.5 & \large 36.06 \small$\pm$ 0.8 & \large 37.74 \small$\pm$ 3.4 & \large 34.52 \small$\pm$ 2.2 & \large 42.88 \small$\pm$ 1.7 & \large 58.94 \small$\pm$ 2.8 & \large 66.12 \small$\pm$ 2.1 & \large 42.41 \small$\pm$ 14.2 & \large \underline{72.36} \small$\pm$ 1.9 & \large 65.41 \small$\pm$ 1.3 & \large 41.11 \small$\pm$ 5.0 & \large \textbf{72.86} \small$\pm$ 0.9\\
        \bottomrule
    \end{tabular}
     }
    \label{tab:expriment}
\end{table*}

\begin{figure*}[!t]
\vspace{-6mm}
\centering
\subfloat[scziDesk]{
\includegraphics[width=0.18\textwidth]{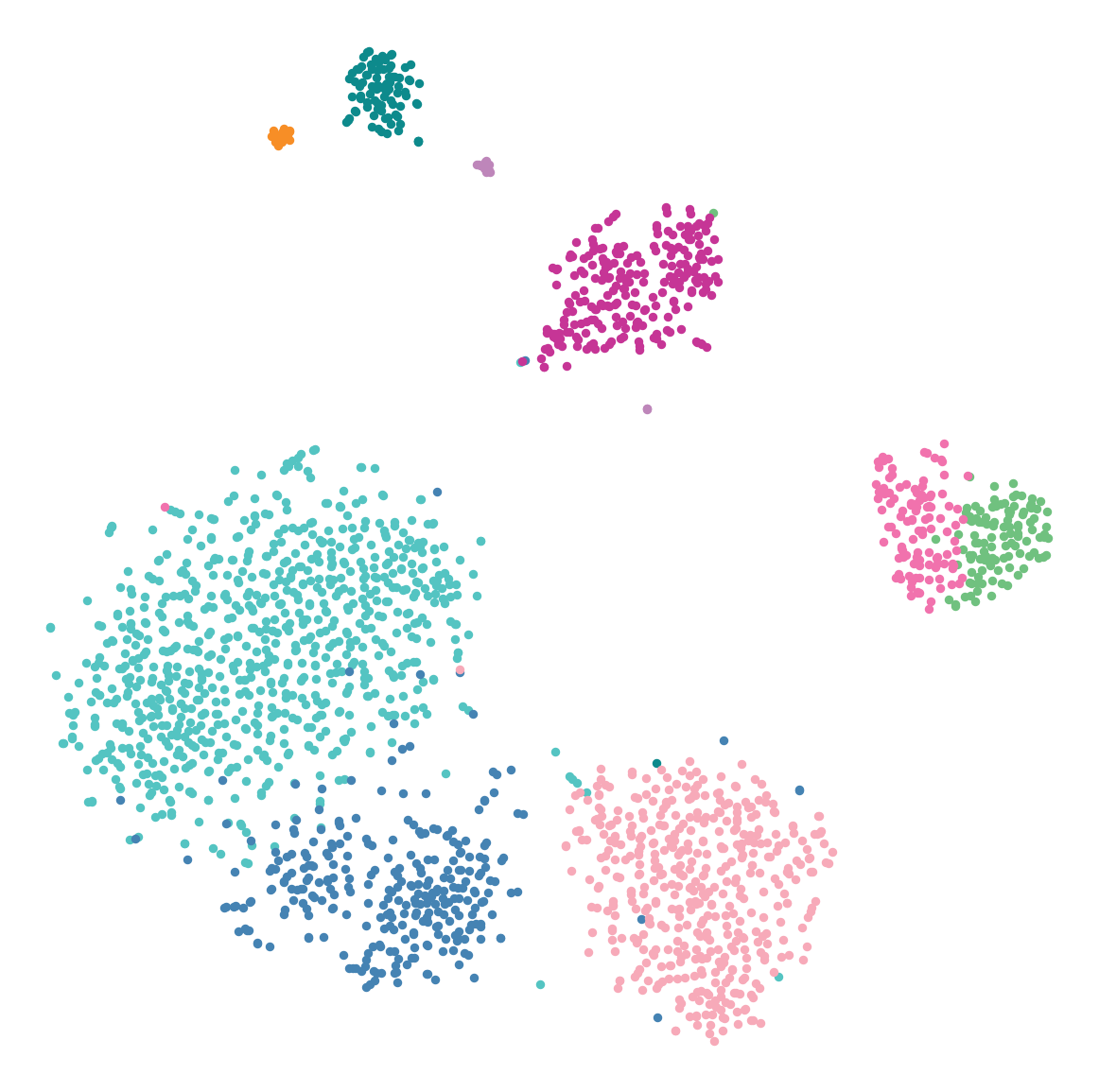}
}
\subfloat[scDeepCluster]{
\includegraphics[width=0.18\textwidth]{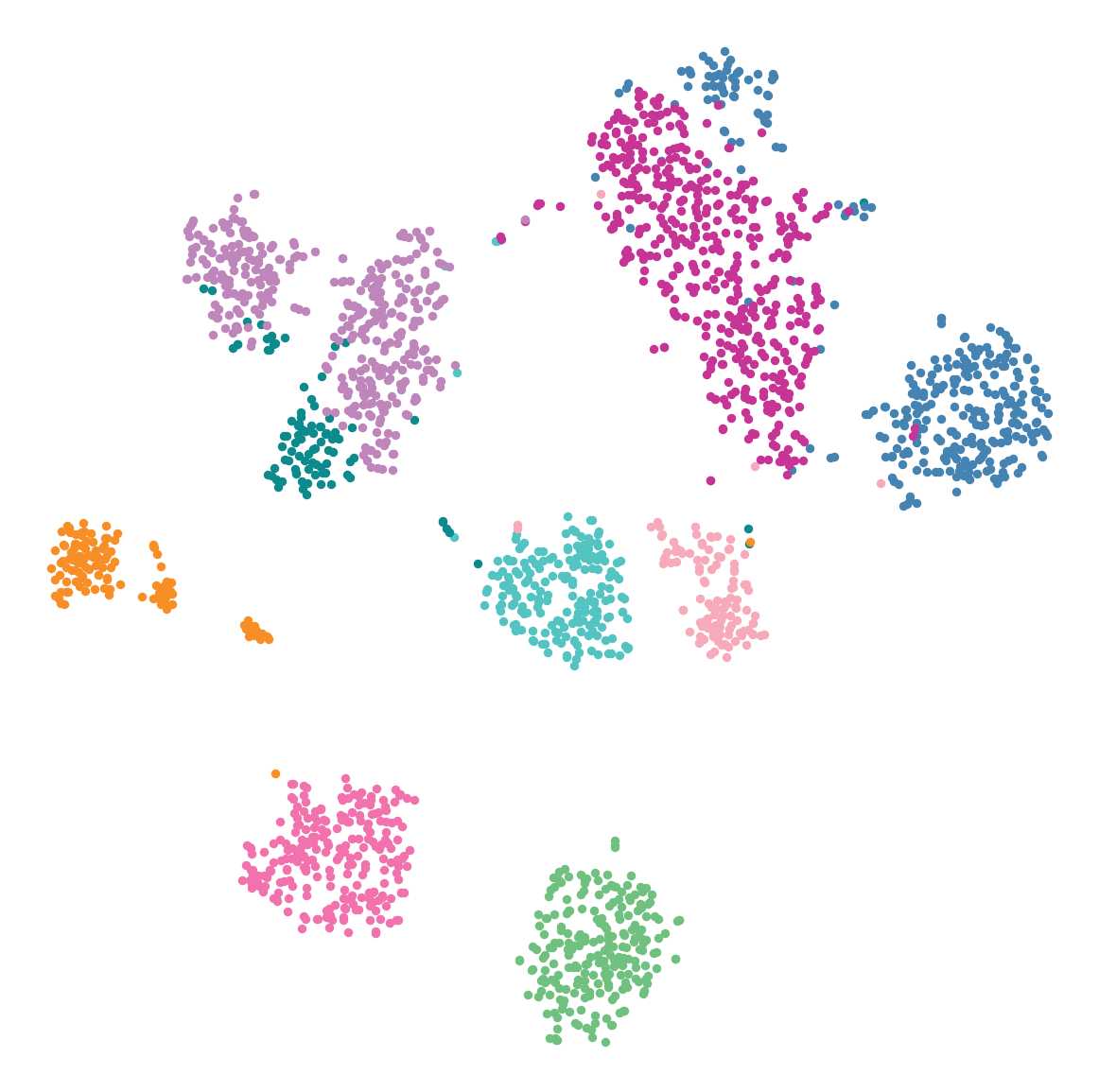}
}
\subfloat[scGNN]{
\includegraphics[width=0.18\textwidth]{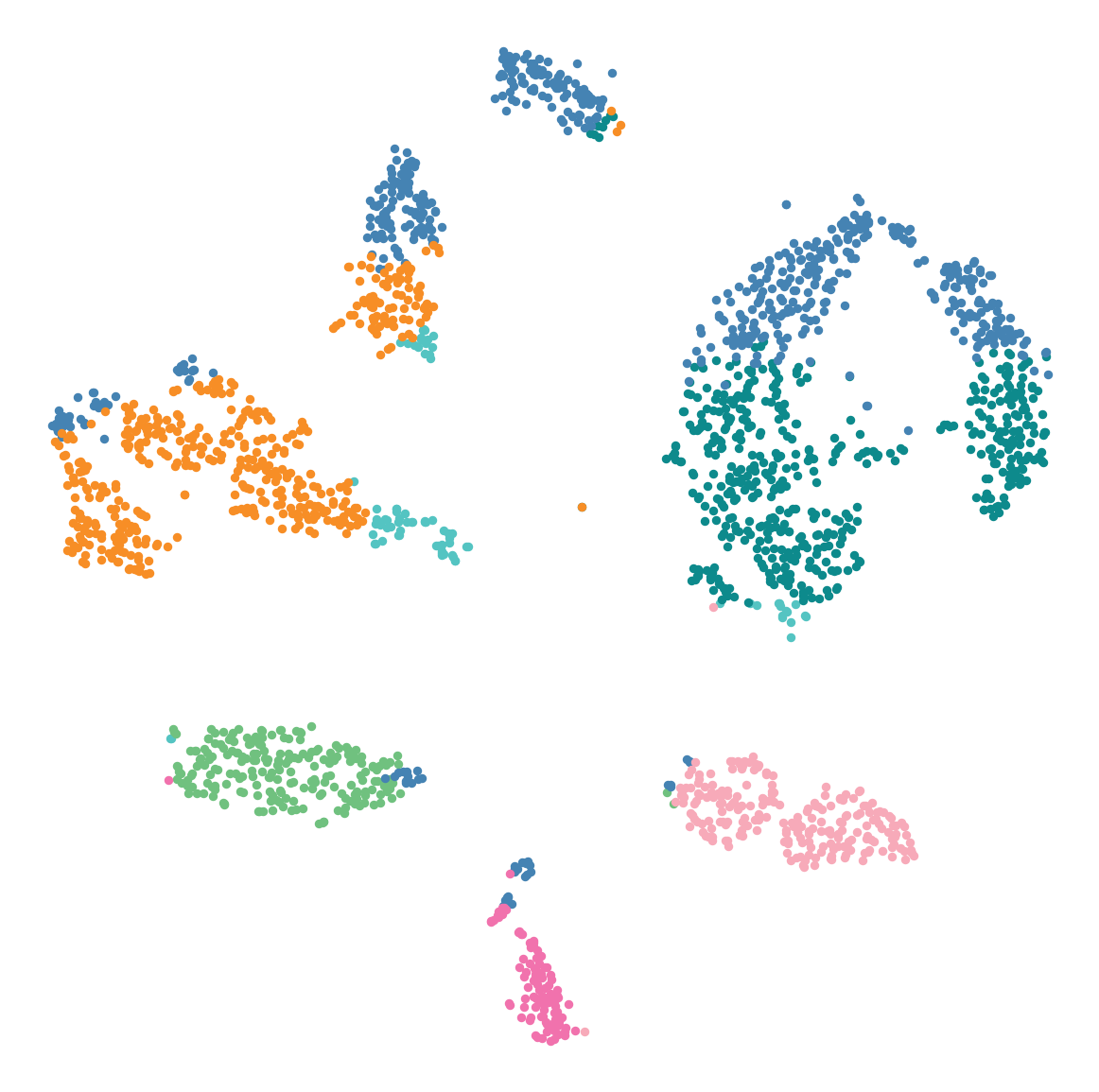}
}
\subfloat[scCDCG]{
\includegraphics[width=0.18\textwidth]{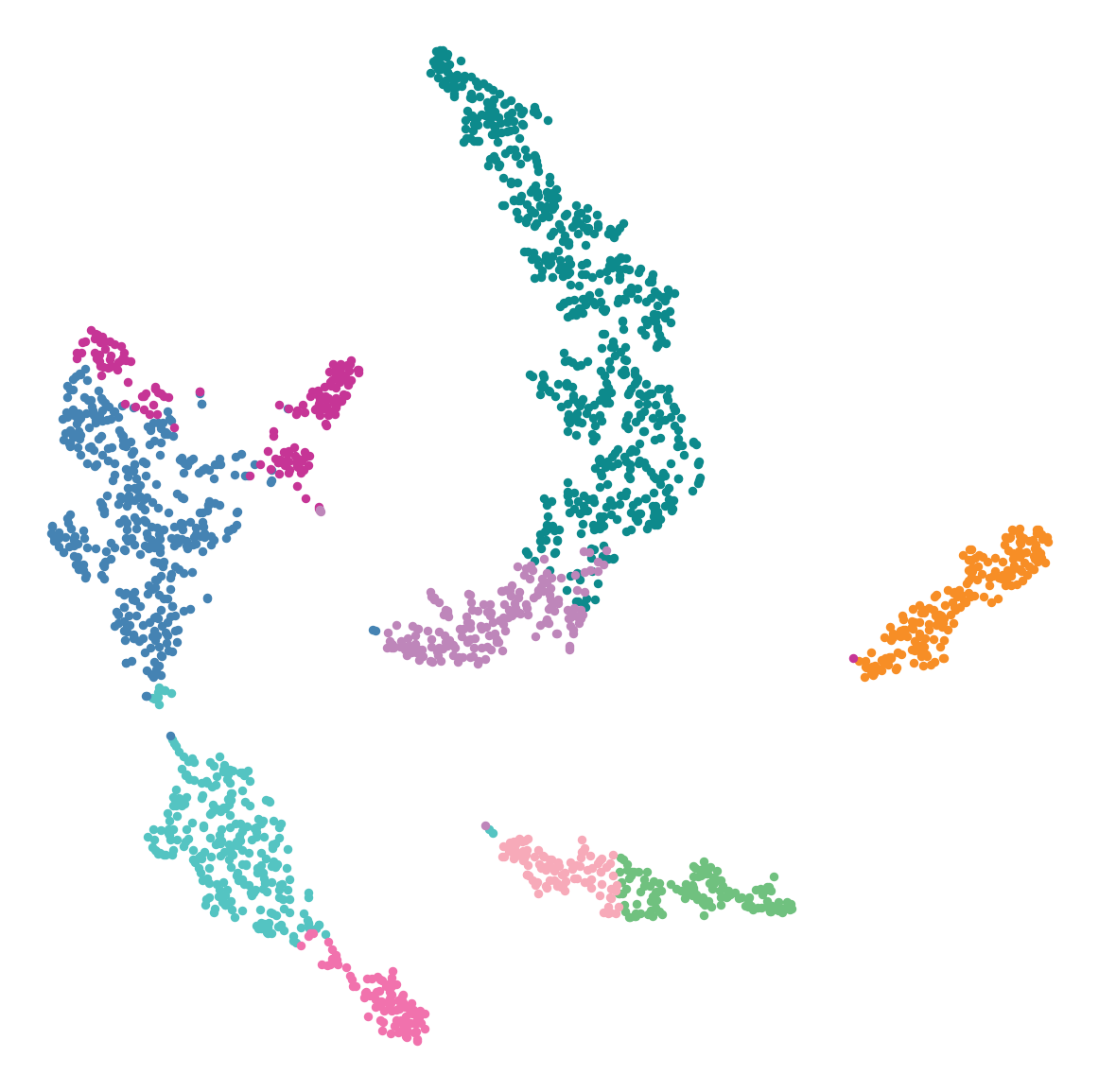}
}
\subfloat[~\methodname~]{
\includegraphics[width=0.18\textwidth]{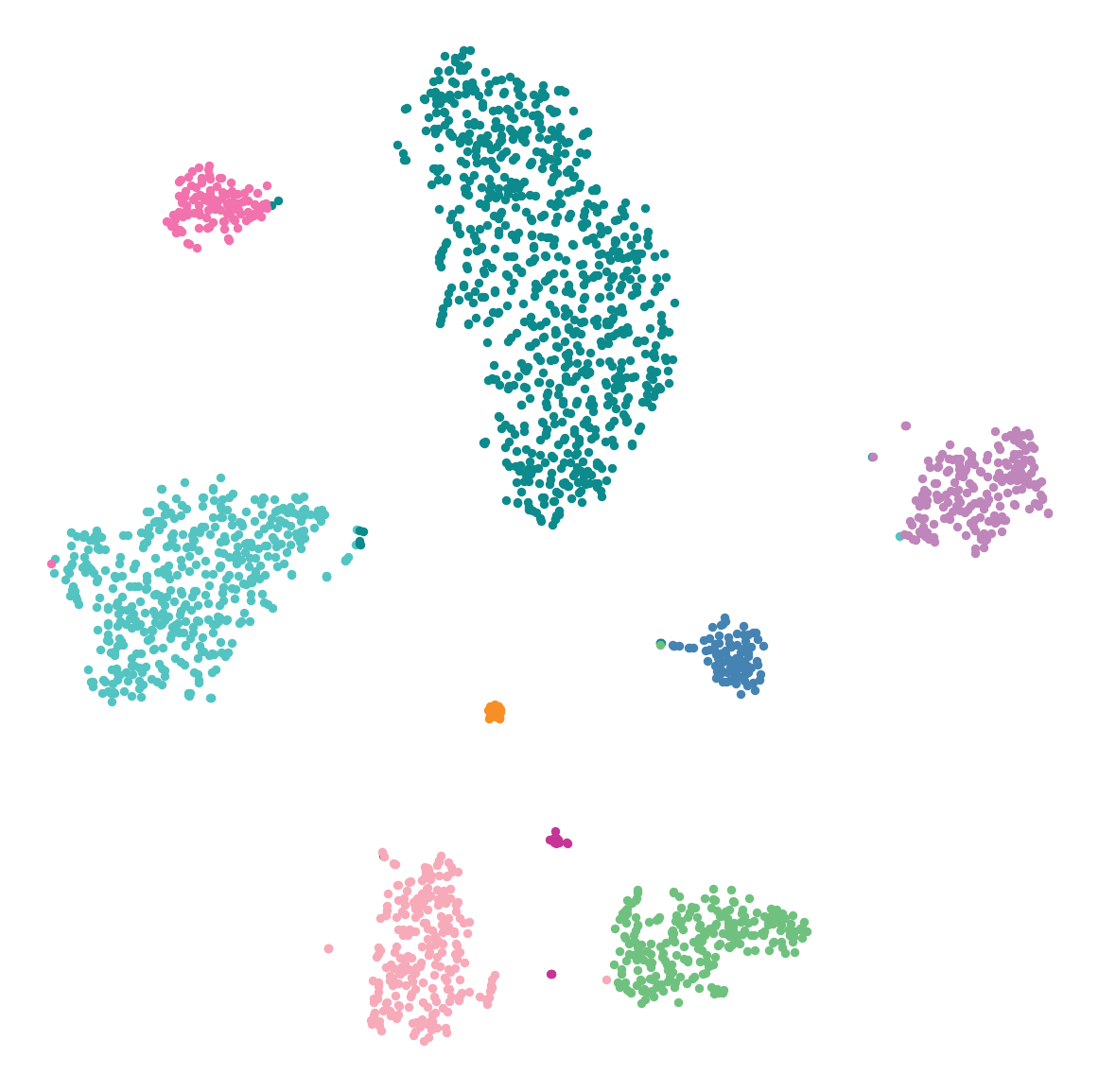}
}
\caption{Visualization of~\methodname~and four typical baselines on~\emph{Mauro Human Pancreas cells} dataset in 2D t-SNE projection. Each point represents a cell, while each color represents a predicted cell type.}
\vspace{-6mm}
\label{fig: visualization}
\end{figure*}

\subsubsection{\textbf{Qualitative Analysis}}
To intuitively interpret the clustering results from a biological perspective, we use t-SNE to extract and visualize the distribution of clustering embeddings on the~\emph{Mauro Human Pancreas cells} dataset. The embeddings, represented by the matrix $\mathbf{Z}$ in a two-dimensional space, are obtained from several classical baseline methods.
As shown in Fig.~\ref{fig: visualization},~\methodname~effectively differentiates cell types by learning compact, well-separated representations that form distinct boundaries between populations, thereby better capturing the underlying clustering structure in scRNA-seq data. 
In contrast, other methods exhibit suboptimal clustering performance, with noticeable overlap between clusters and more scattered results, failing to group similar cell types effectively.


\subsubsection{\textbf{Similarity Analysis}} 
\begin{figure*}[t!]
\centering
\subfloat[DEC]{
\includegraphics[width=0.25\textwidth]{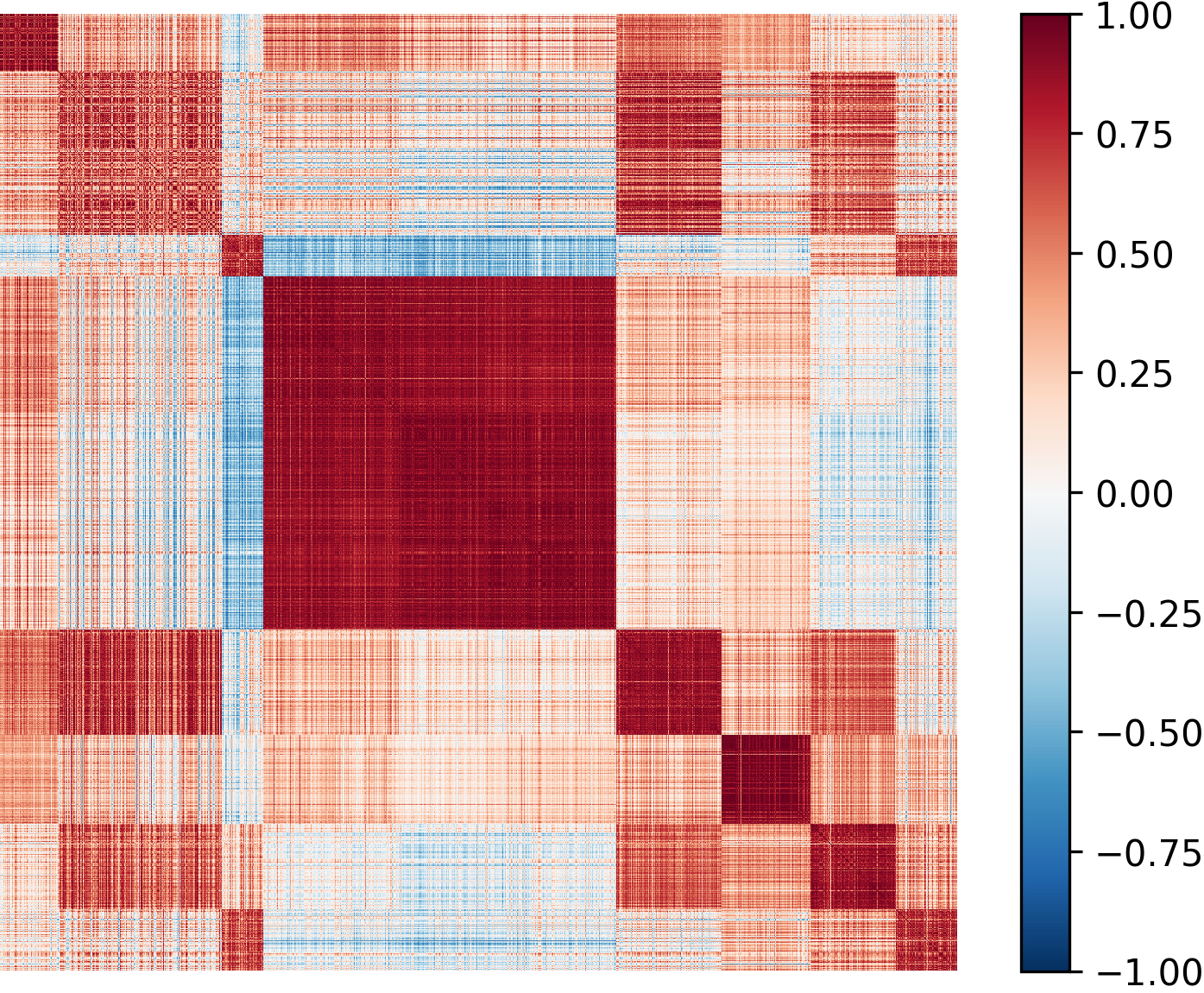}
}
\subfloat[scDeepCluster]{
\includegraphics[width=0.25\textwidth]{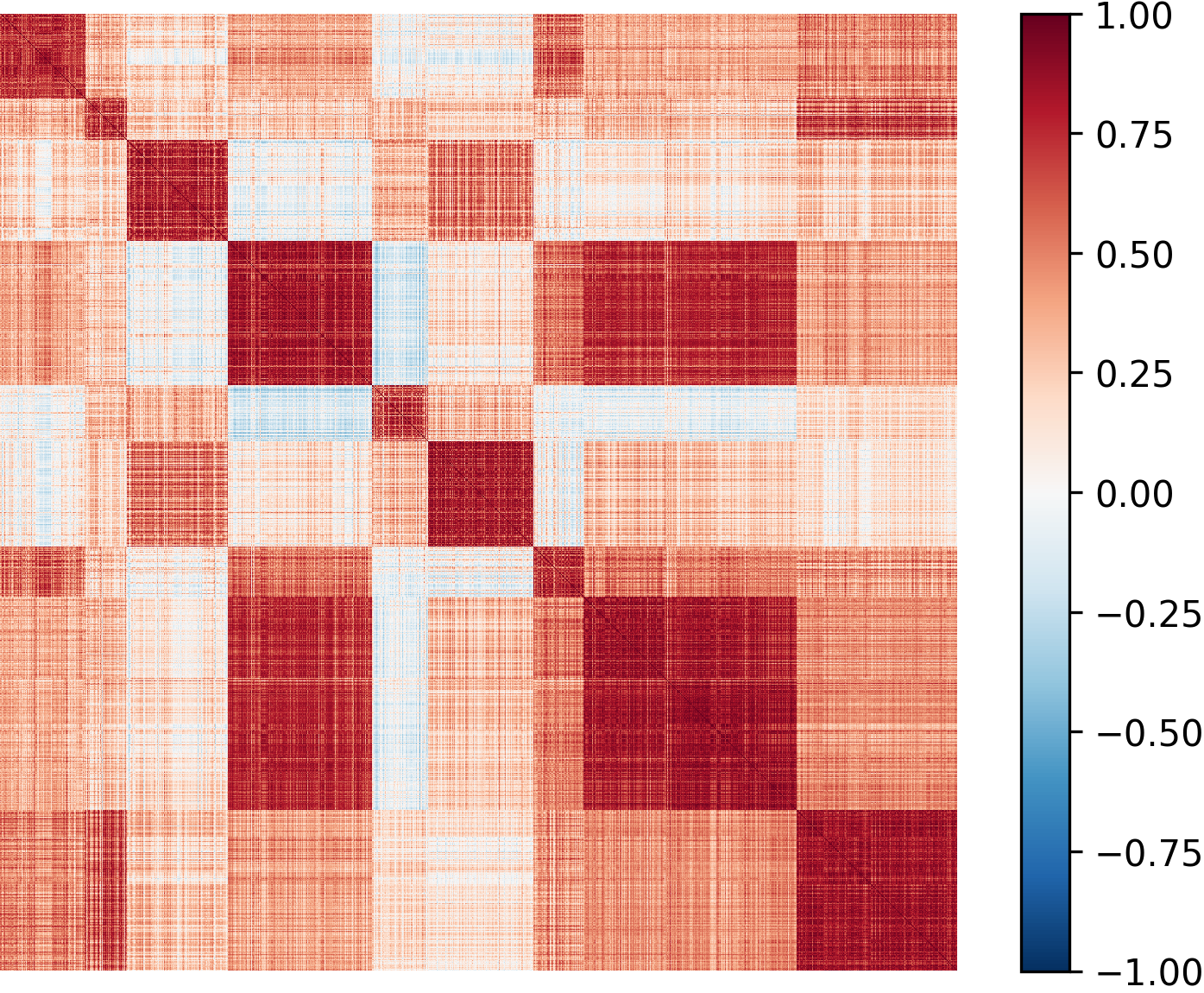}
}
\subfloat[scCDCG]{
\includegraphics[width=0.25\textwidth]{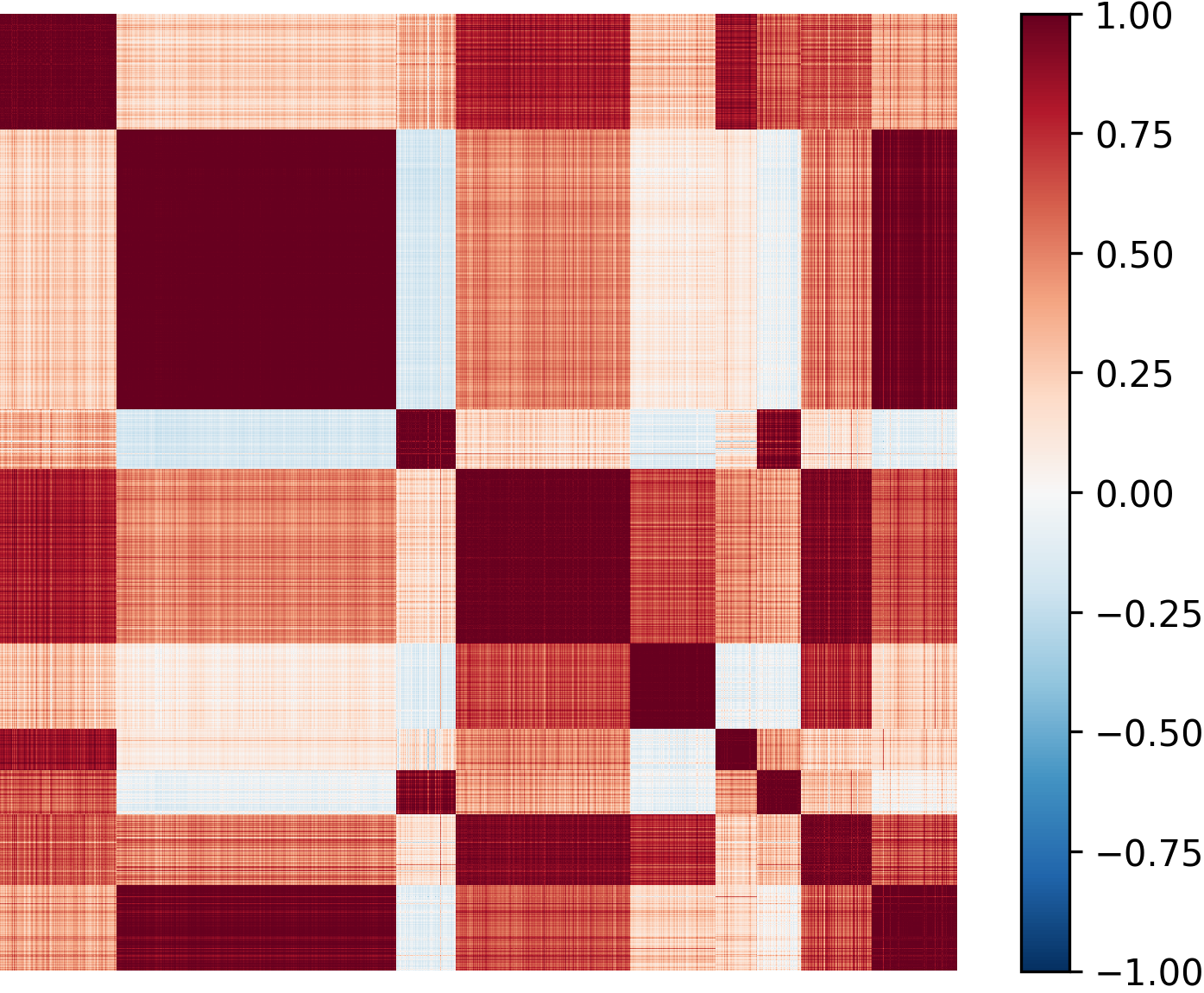}
}
\subfloat[\methodname]{
\includegraphics[width=0.25\textwidth]{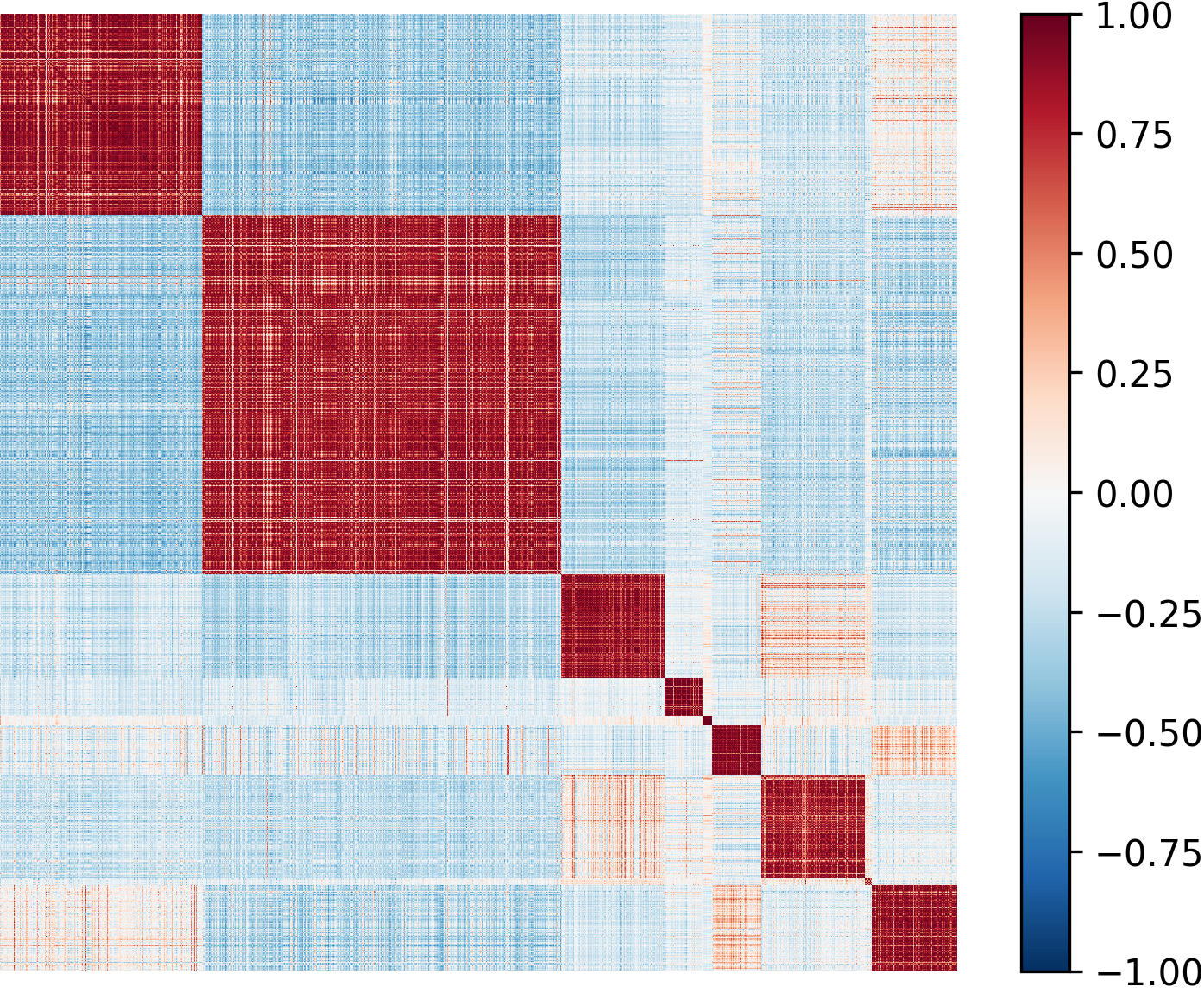}
}
\caption{The heat maps of node similarity matrices in the latent space of DEG, scDeepCluster, scGNN, and our proposed method~\methodname~on the~\emph{Mauro Human Pancreas cells} dataset.}
\label{fig: fig1_heatmap}
\vspace{-3mm}
\end{figure*}

Taking the~\emph{Mauro Human Pancreas cells} dataset as an example, we extract representation matrices from~\methodname~and three representative methods, construct their element similarity matrices, and visualize the results as heatmaps.
As shown in Fig.~\ref{fig: fig1_heatmap},~\methodname~effectively captures the three-dimensional clustering structure in the latent space, forming more compact and well-organized clusters, which indicates its advantage in preserving continuous cell similarities and revealing the underlying biological structure. 
In contrast, the structural information of other methods is more scattered, making it difficult to accurately characterize potential biological features.
This comparison further demonstrates the superiority of~\methodname~in enhancing intercellular similarity representation and maintaining structural integrity.

\subsection{\textbf{Biological Analysis}}
Following the validation of the clustering performance of~\methodname, it is essential to further interpret its biological significance. 
Clustering single-cell RNA sequencing data enables the identification of gene expression patterns, offering valuable insights into cellular heterogeneity. 
Such analysis is crucial for understanding the tissue microenvironment and advancing precision medicine applications. 
In this section, we use the~\emph{Mauro Human Pancreas cells} dataset as a case study to perform biological validation based on the clustering labels generated by~\methodname. 
Specifically, we conduct marker gene identification and cell type annotation to clarify the biological relevance of the clustering results.

\begin{figure*}[!t]
    \centering
    \includegraphics[width=0.9\linewidth]{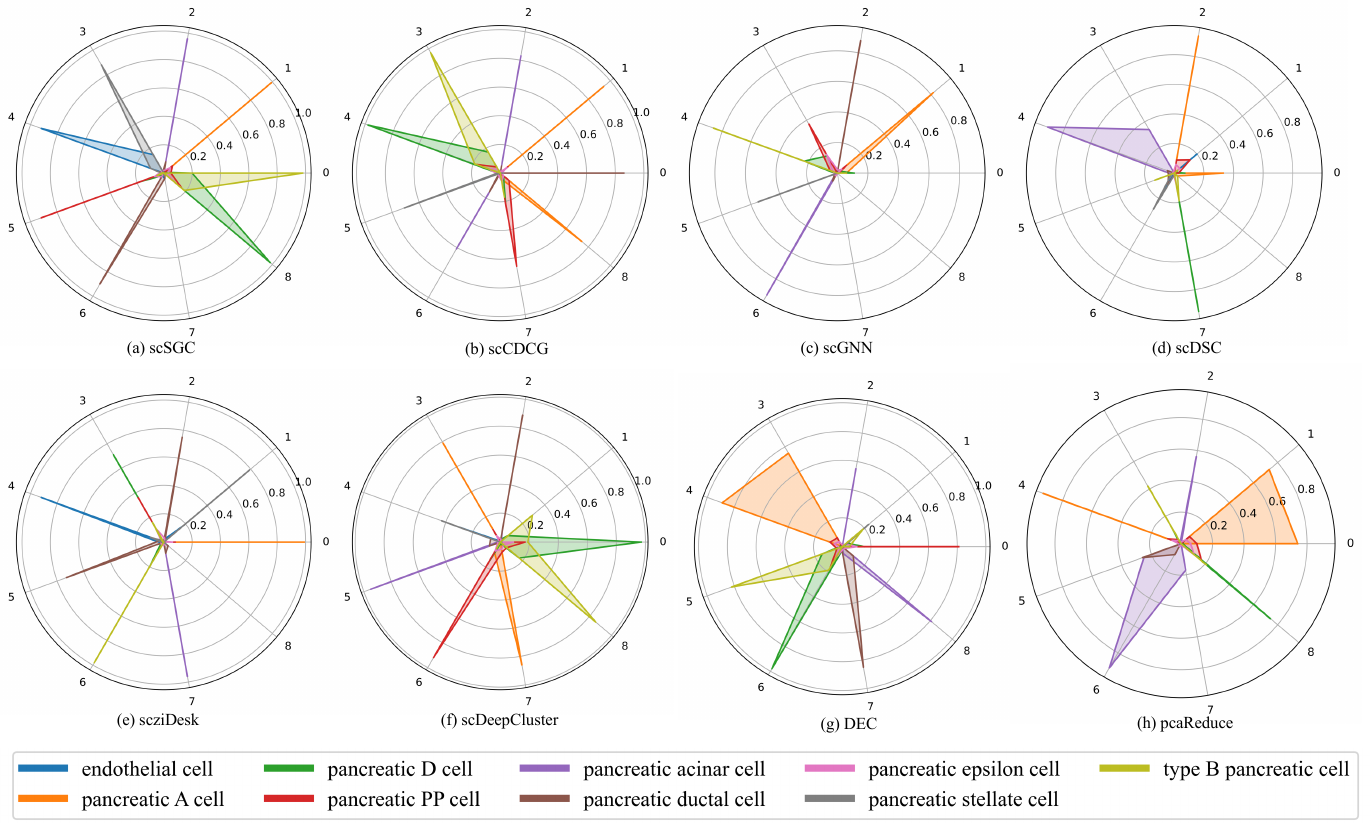}
    \caption{Cell type annotation. Overlap of top 100 DEGs in clusters detected by ten methods with gold standard cell types (similarity = number of overlapped DEGs/100).}
    \vspace{-5mm}
    \label{fig:cell_type_annotation}
\end{figure*}

\subsubsection{\textbf{Cell Type Annotation}}
Differentially expressed genes (DEGs) serve as a critical indicator for cell type annotation, as they allow the marker gene identification within each cluster based on gene expression matrices and predicted labels. 
We employ the ``FindAllMarkers" function from the Seurat~\cite{butler2018integrating} package to identify the DEGs for each cluster. 
Taking the~\emph{Mauro Human Pancreas cells} as an example, we first obtained the top 100 DEGs within gold standard clusters for each annotated cell type, which served as a baseline for evaluating various methods. 
We apply the same approach to both~\methodname~and other competing methods, identifying the top 100 marker genes for each cluster and calculating the overlap with the gold standard DEGs, as illustrated in Fig.~\ref{fig:cell_type_annotation}. 
\methodname~accurately assigned cell types to specific clusters, with DEGs overlap exceeding 90\% for all clusters except cluster 7. 
For example, clusters 2, 1, 0, 8, 6, 4, 5, and 3 were annotated as `pancreatic acinar cell', `pancreatic A cell', `type B pancreatic cell', `pancreatic D cell', `pancreatic ductal cell', `endothelial cell', `pancreatic PP cell', and `pancreatic stellate cell', respectively. 
The corresponding sample sizes for these eight cell types in the gold standard were 219, 812, 448, 193, 245, 21, 101, and 80, while `pancreatic epsilon cell' had only 3 samples. 
Due to the extremely small sample size of 'pancreatic epsilon cell' in the gold standard,~\methodname~did not provide a clear annotation for this type but successfully identified the other eight cell types, demonstrating high accuracy and reliability. 

Fig.~\ref{fig:cell_type_annotation}(b)-(h) presents the cell type annotation results of seven other methods. 
Specifically, scGNN failed to identify the cell types of clusters 0, 7, and 8, scDeepCluster could not accurately partition cluster 1, and the DEG overlap for cluster 4 was below 40\%. 
DEC misannotated clusters 3 and 4 as 'pancreatic A cell'. 
This demonstrates~\methodname's superior ability to accurately match clusters to known cell types, highlighting the lack of consistency and accuracy in competing methods.

\subsubsection{\textbf{Marker Gene Identification}}
\begin{figure*}[t!]
    \centering
    \includegraphics[width=0.9\linewidth]{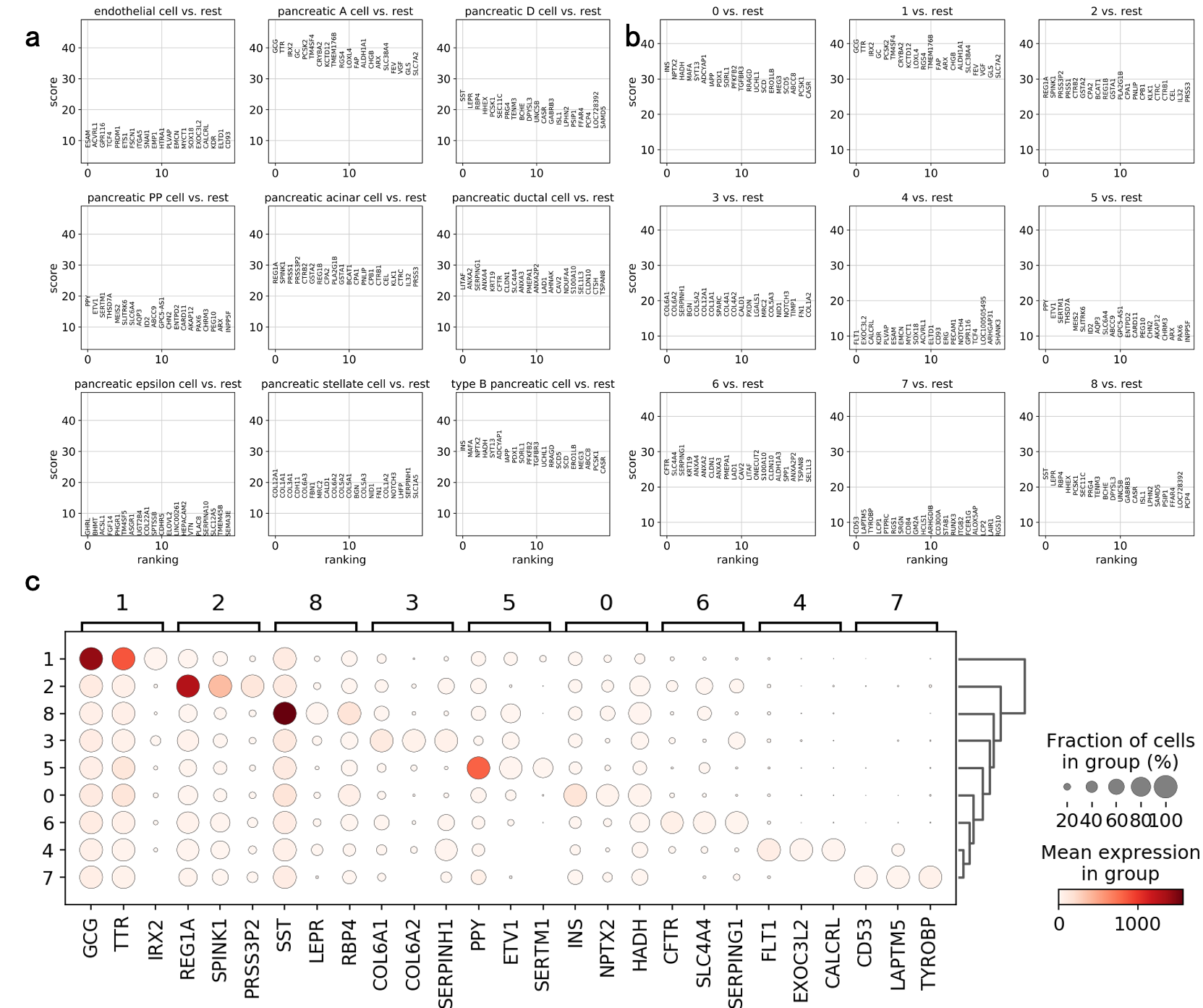}
    \caption{Marker gene identification. (a) Top 20 DEGs in different cell types by gold standard. (b) Top 20 DEGs in different cell types by~\methodname. (c) Dot plot of the top 3 DEGs within each cell type by~\methodname.}
    \label{fig:marker_gene}
\end{figure*}
Additionally, we conducted DEGs analysis and marker gene identification based on clusters identified by~\methodname. 
Using the Wilcoxon Rank Sum test, Fig.~\ref{fig:marker_gene}(a) and (b) present the top 20 DEGs for different cell types identified by~\methodname~and the gold standard. 
This allows us to see which genes overlap between the clusters identified by~\methodname~and the gold standard, further supporting the accuracy of the cell type annotations. 
Fig.~\ref{fig:marker_gene}(c) illustrates a dot plot of the top three DEGs for each of the nine clusters. 
These genes represent the most characteristic DEGs identified by~\methodname~for each cluster and can be regarded as marker genes. 
For instance, the marker genes for cluster 1 are GCG, TTR, and IRX2, which serve as defining characteristics of the cell type in this cluster.

\subsection{\textbf{Ablation Study}}
\begin{figure*}[t!]
\centering
\hspace{-9mm}
\subfloat[Huamn Liver cells]{
\includegraphics[width=0.24\textwidth]{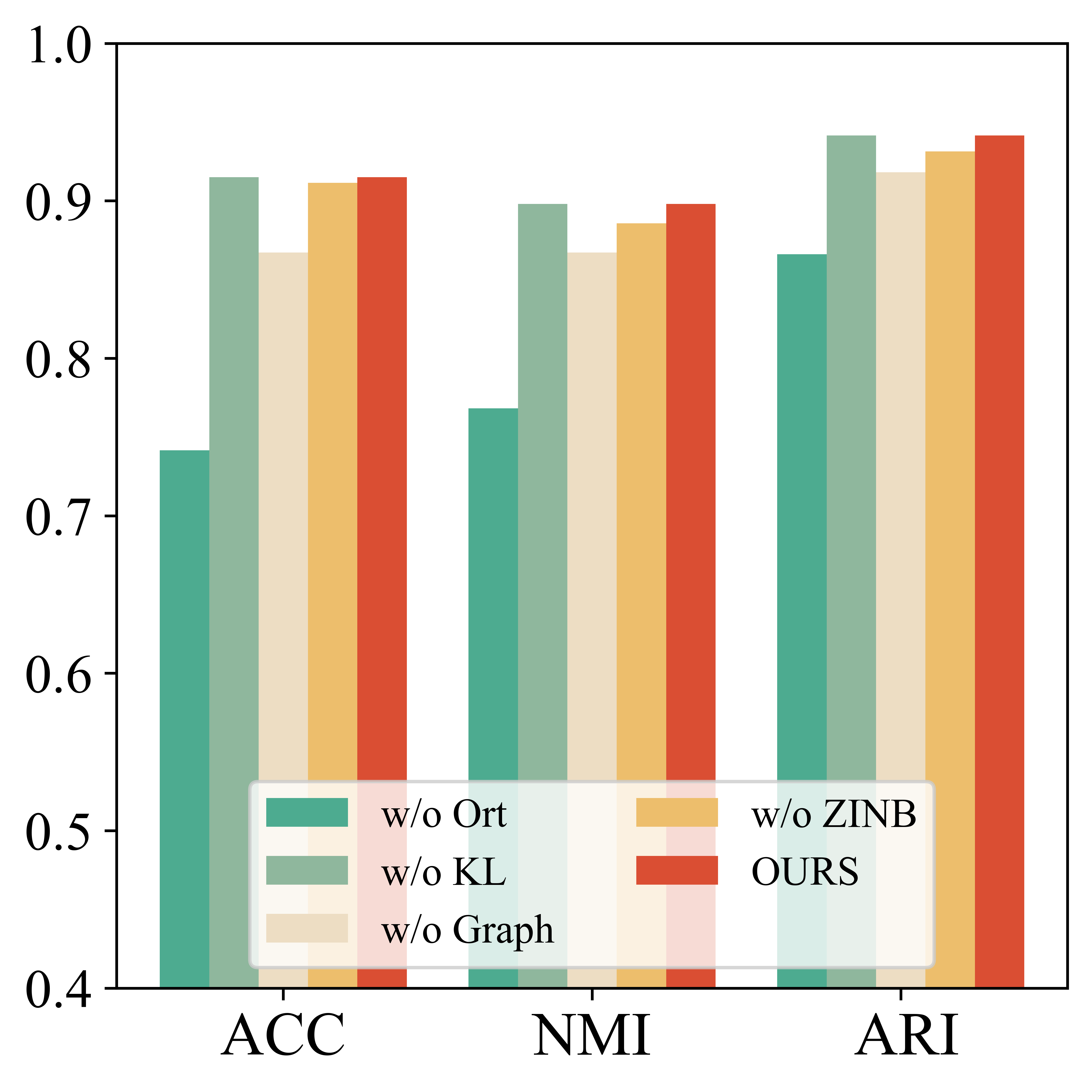}
}
\subfloat[Mauro Human Pancreas cells]{
\includegraphics[width=0.24\textwidth]{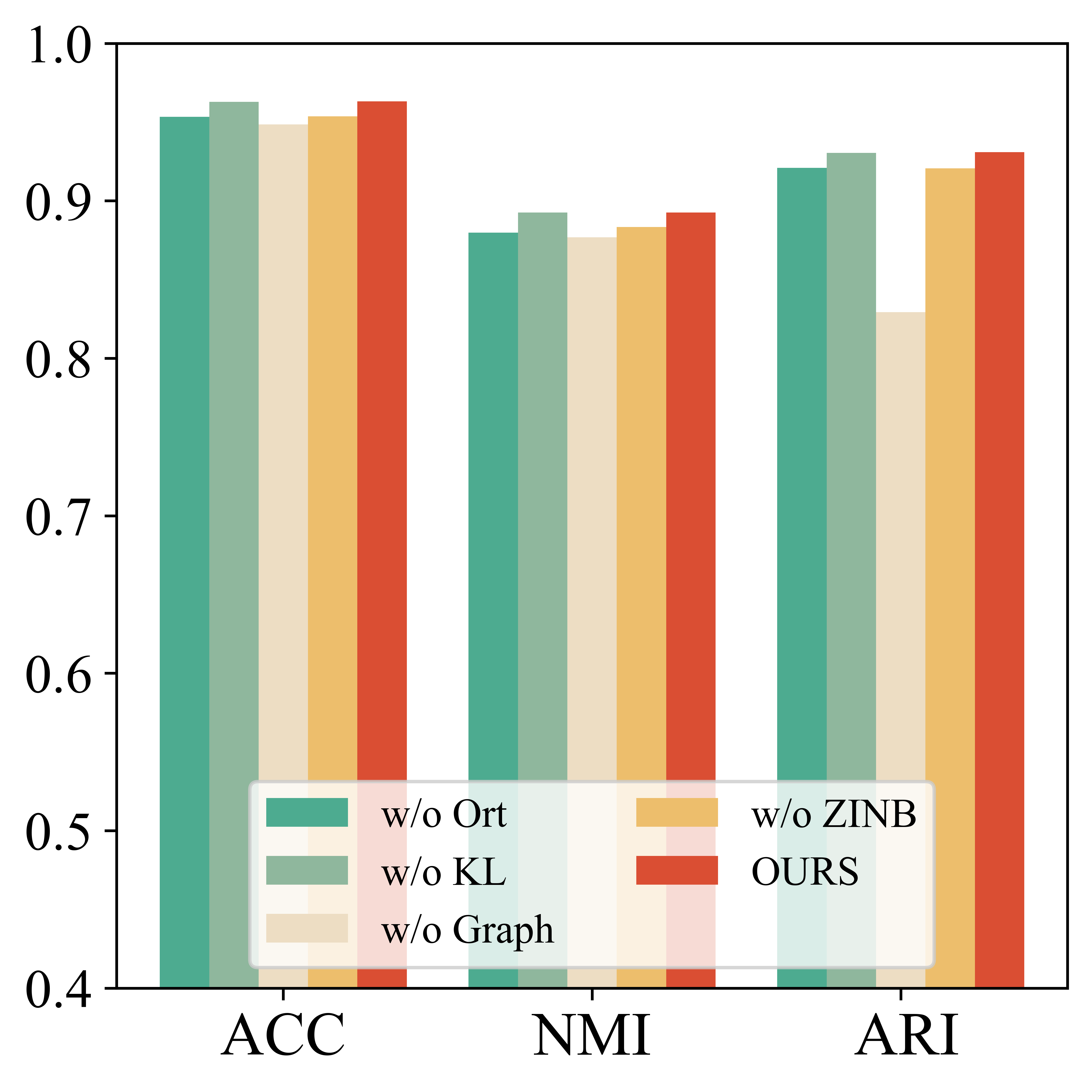}
}
\subfloat[Mouse Pancreas cells 1]{
\includegraphics[width=0.24\textwidth]{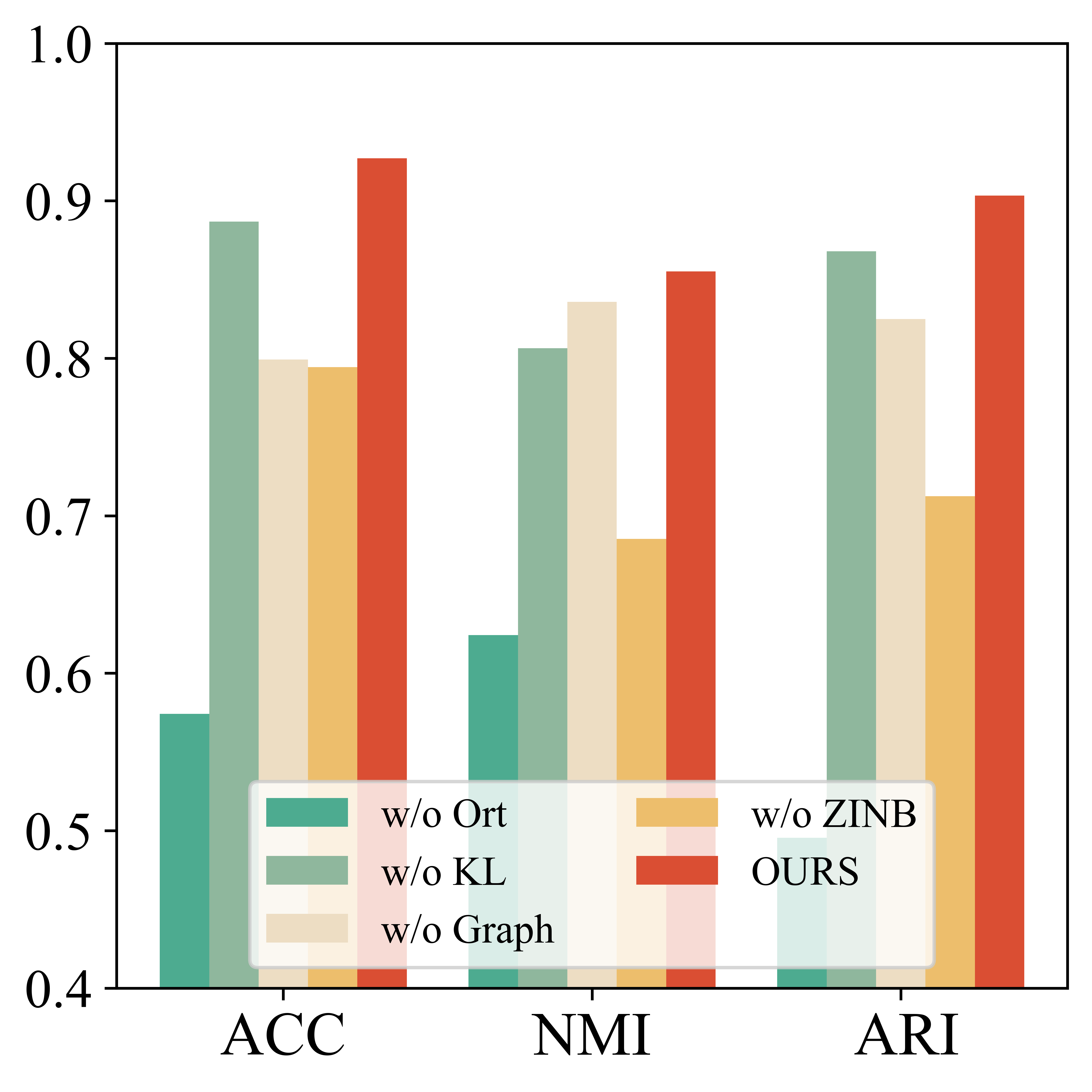}
}
\subfloat[Human Pancreas cells 2]{
\includegraphics[width=0.24\textwidth]{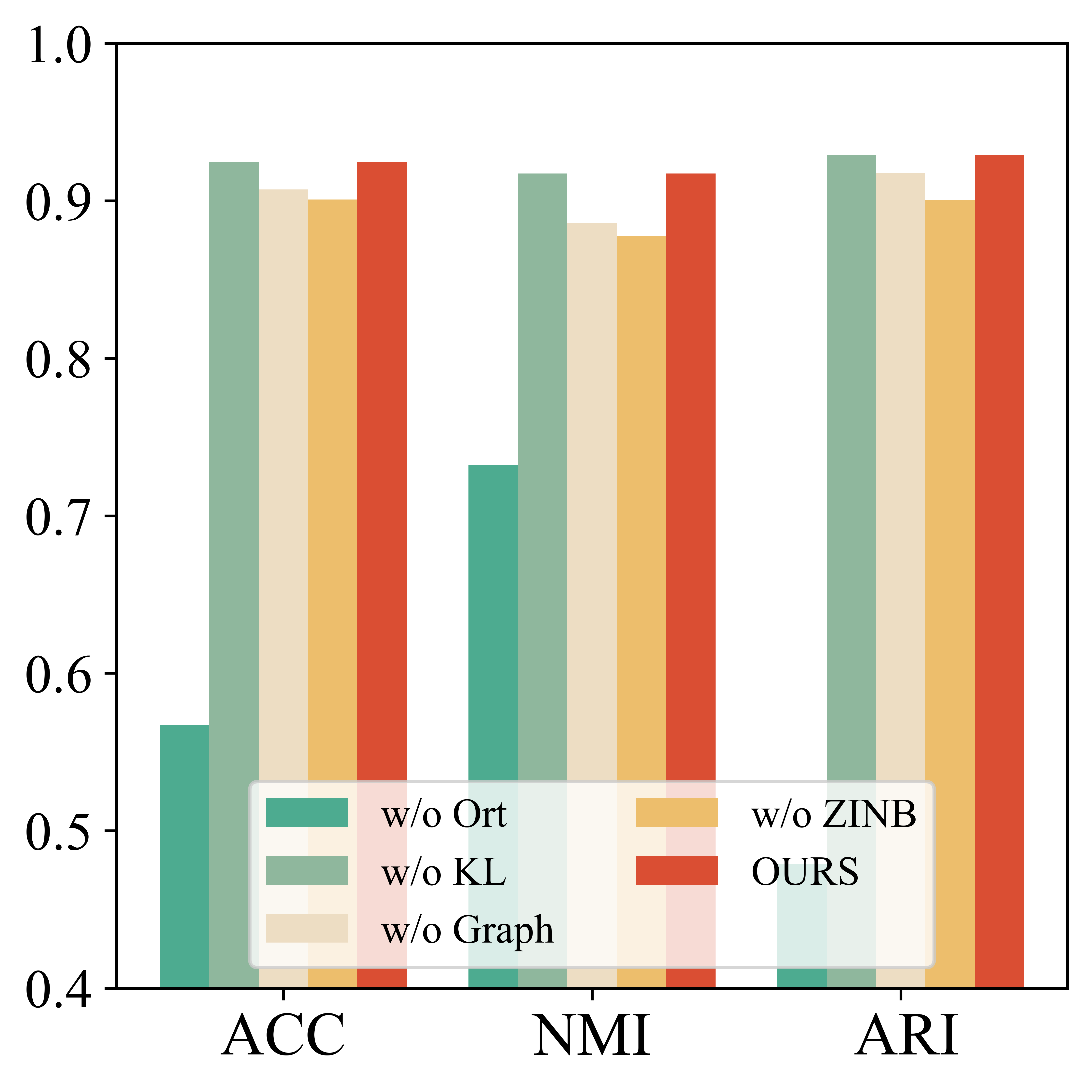}
}
\hspace{-5mm}
\captionsetup{justification=raggedright, singlelinecheck=false} 
\caption{Effectiveness of each main components of ~\methodname.}
\label{fig:ablation of all component}
\end{figure*}
\begin{figure*}[t!]
\centering
\hspace{-9mm}
\subfloat[Huamn Liver cells]{
\includegraphics[width=0.24\textwidth]{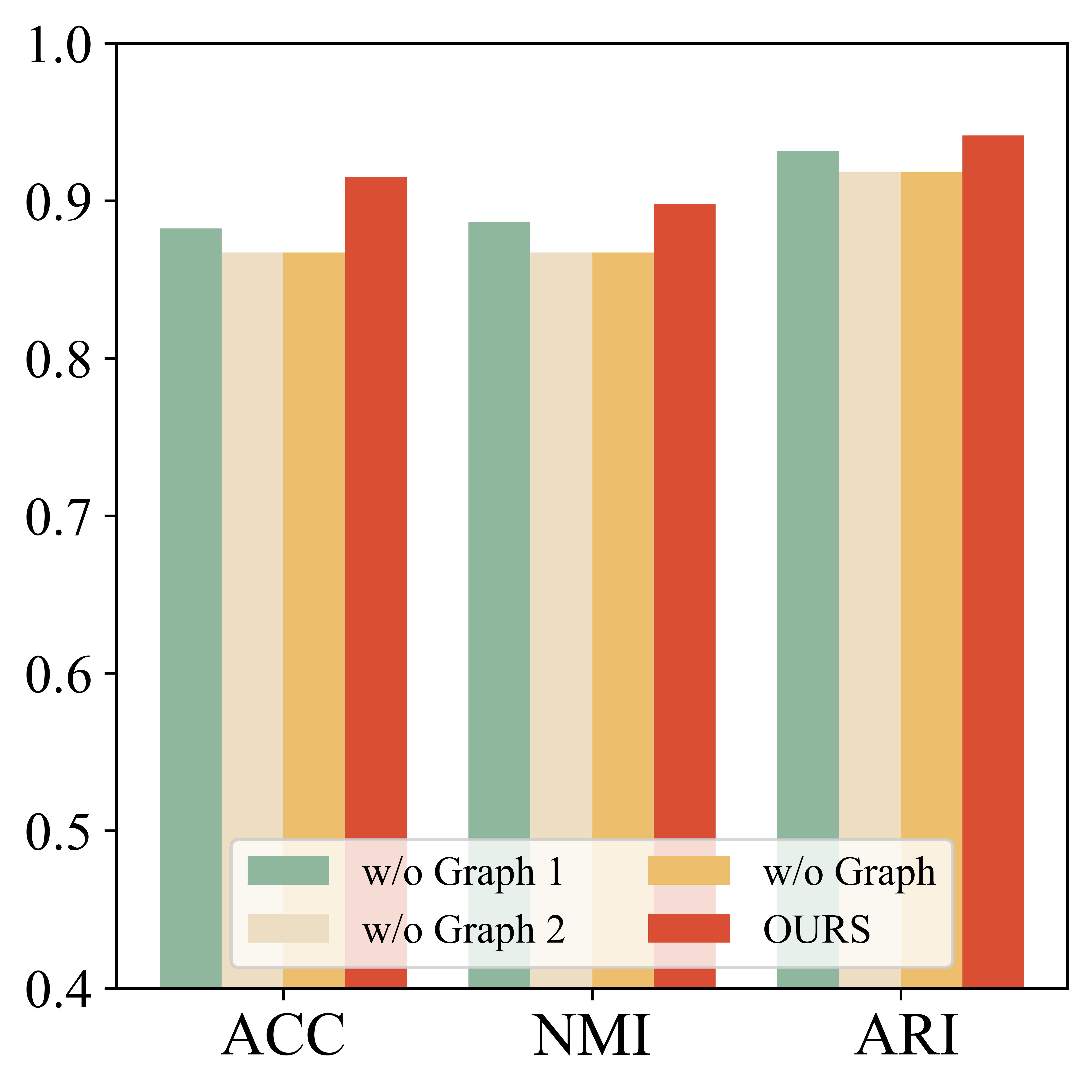}
}
\subfloat[Mauro Human Pancreas cells]{
\includegraphics[width=0.24\textwidth]{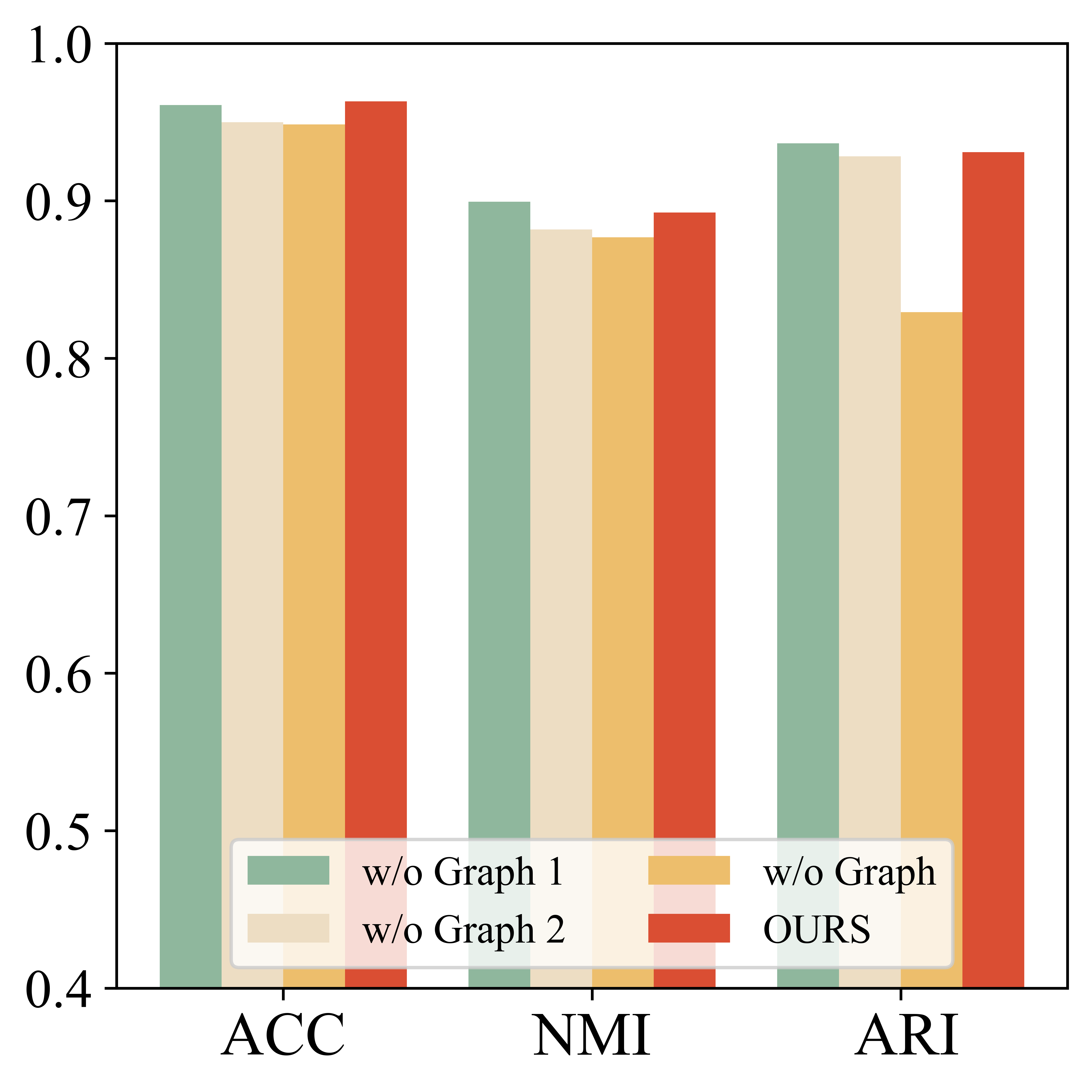}
}
\subfloat[Mouse Pancreas cells 1]{
\includegraphics[width=0.24\textwidth]{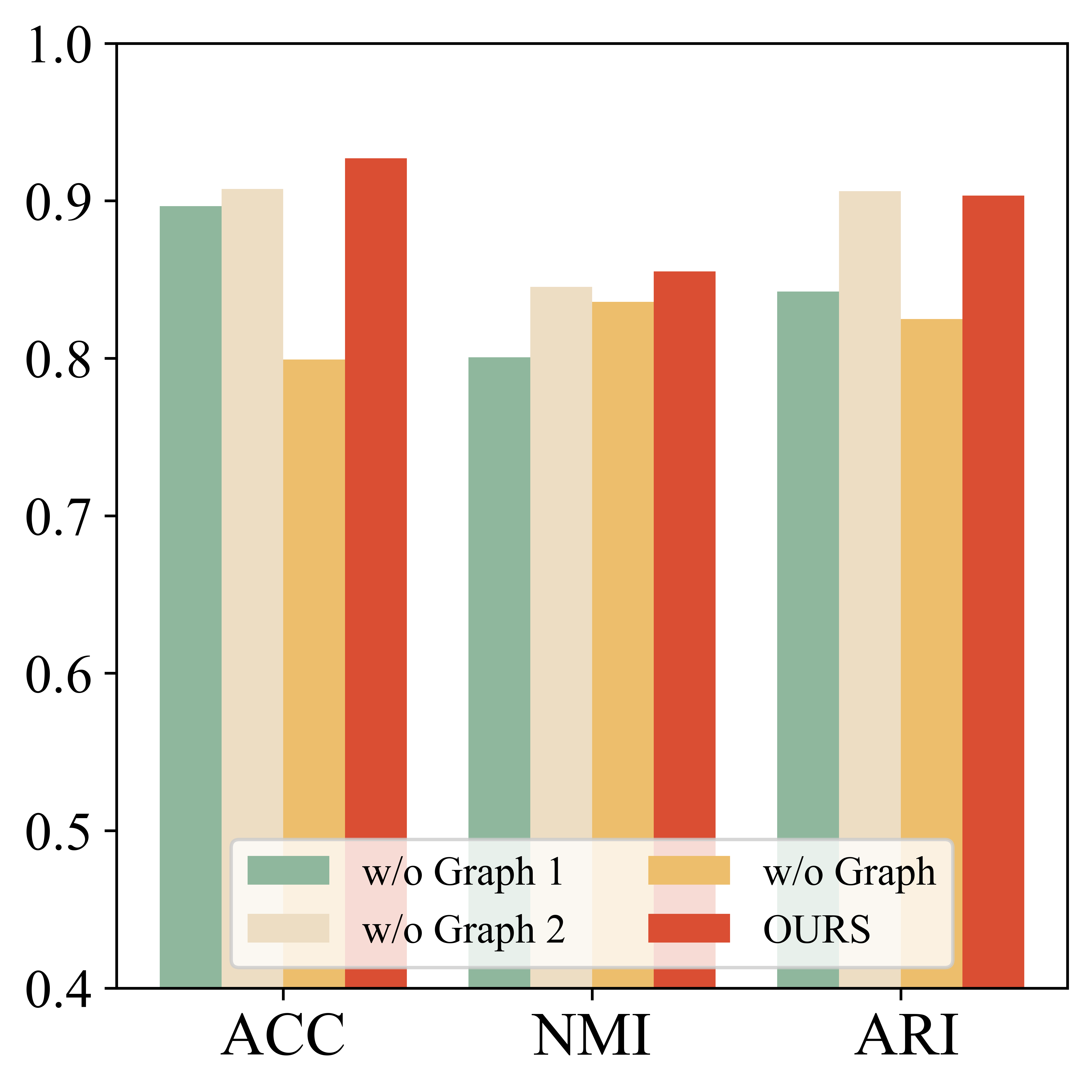}
}
\subfloat[Human Pancreas cells 2]{
\includegraphics[width=0.24\textwidth]{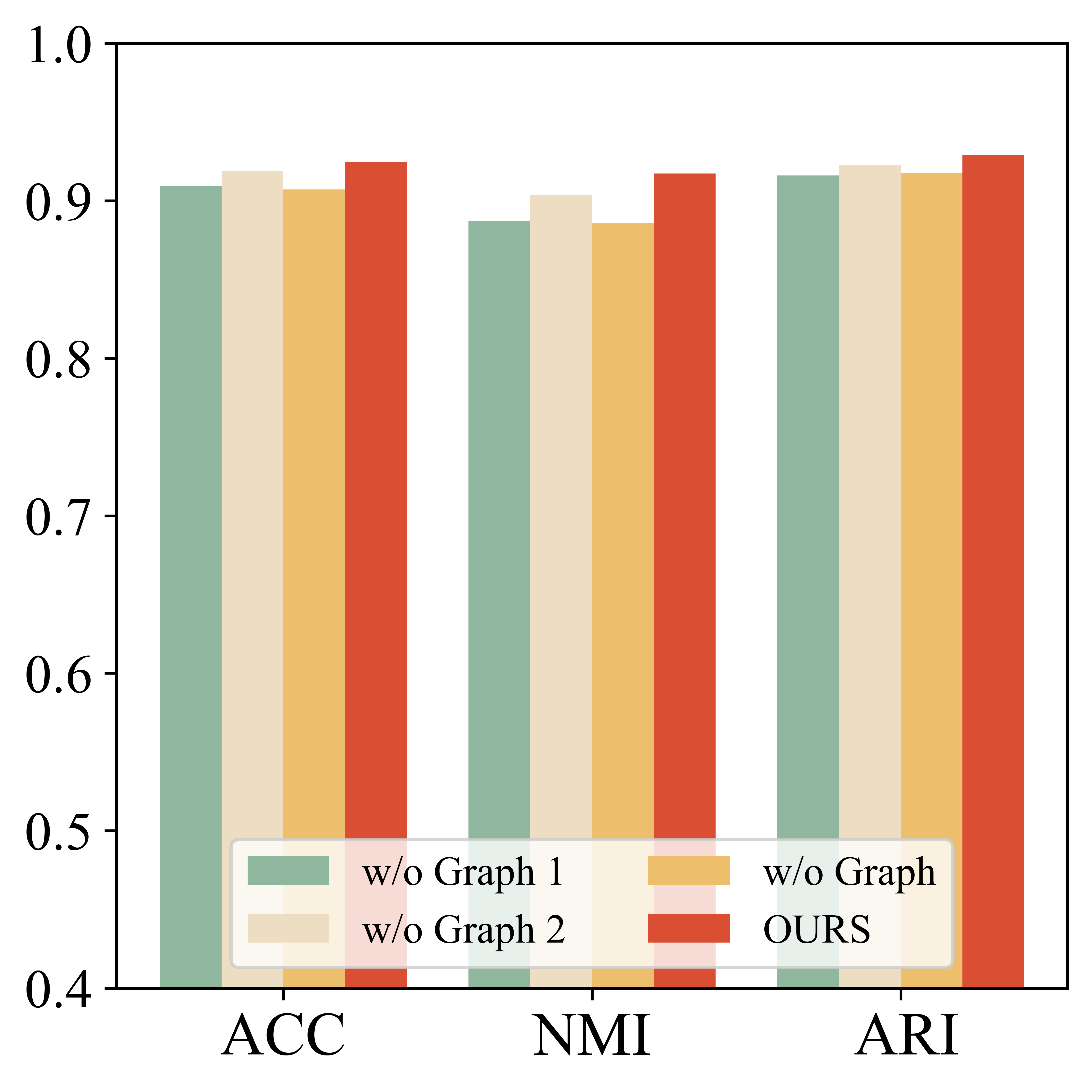}
}
\hspace{-5mm}
\captionsetup{justification=raggedright, singlelinecheck=false} 
\caption{Effectiveness of each component of graph information on model performance.}
\label{fig:ablation of graph}
\end{figure*}

\subsubsection{\textbf{Effectiveness of each main components}}
To validate the effectiveness of each component in ~\methodname, we performed ablation studies on four datasets, assessing the contributions of the three main modules related to the NCut loss, ZINB loss, and clustering loss. 
The NCut loss includes two sub-losses: the graph structure information and the orthogonality term.

Here,~\methodname~without the ZINB loss is denoted as~\emph{scSGC-w/o ZINB}, while~\methodname~without KL loss is referred to as~\emph{scSGC-w/o KL}. Furthermore,~\emph{scSGC-w/o Graph} and~\emph{scSGC-w/o Ort} represent models that exclude the graph information and orthogonality components of NCut loss, respectively. The complete model, incorporating all components, is referred to as~\emph{OURS}, i.e.,~\methodname. 
As demonstrated in Fig.~\ref{fig:ablation of all component}, the systematic ablation study reveals that each components of~\methodname~contribute synergistically to performance optimization. 
Compared to other variations, these components significantly improve~\methodname's efficiency and robustness.


\begin{table}[t!]
\centering
\scriptsize
\tiny
\setlength{\tabcolsep}{1pt}  
\renewcommand{\arraystretch}{1.1}  
\caption{Ablation comparisons of optimal transport on two datasets (mean $\pm$ standard deviation).}
\begin{tabular}{lccc}
\toprule
\textbf{\large Dataset} & \textbf{\large Metric} & \textbf{\emph{\large DEC}} & \textbf{\emph{\large OURS}}\\
\midrule
\multirow{3}*{\large Human liver cells} 
    & \large ACC & \large 76.87\textsubscript{$\scriptscriptstyle\pm$1.3} & \large 91.16\textsubscript{$\scriptscriptstyle\pm$0.3} \\
    & \large NMI & \large 72.74\textsubscript{$\scriptscriptstyle\pm$2.5} & \large 89.17\textsubscript{$\scriptscriptstyle\pm$0.6} \\
    & \large ARI & \large 83.47\textsubscript{$\scriptscriptstyle\pm$2.2} & \large 93.65\textsubscript{$\scriptscriptstyle\pm$0.5} \\
\midrule
\multirow{3}*{\large Mauro Human Pancreas cells} 
    & \large ACC & \large 88.52\textsubscript{$\scriptscriptstyle\pm$2.5} & \large 96.03\textsubscript{$\scriptscriptstyle\pm$0.2} \\
    & \large NMI & \large 83.68\textsubscript{$\scriptscriptstyle\pm$2.0} & \large 90.23\textsubscript{$\scriptscriptstyle\pm$1.7} \\
    & \large ARI & \large 86.19\textsubscript{$\scriptscriptstyle\pm$5.4} & \large 92.39\textsubscript{$\scriptscriptstyle\pm$1.9} \\
\bottomrule
\end{tabular}
\label{tab:ablation_opt}
\end{table}

\subsubsection{\textbf{Effectiveness of each component of graph information}}
As shown in Fig.~\ref{fig:ablation of all component} , the incorporation of soft graph information significantly contributes to the overall performance improvement. 
To further validate the importance of different components within the soft graph, we conducted experiments comparing \methodname (denoted as~\emph{OURS}) with its various ablated versions. 
Specifically,~\emph{scSGC-w/o Graph 1} and~\emph{scSGC-w/o Graph 2} represent the models without the feature similarity graph $\mathcal{G}_1$ and the cosine similarity graph $\mathcal{G}_2$, respectively. 
Notably, Fig.~\ref{fig:ablation of graph} demonstrates that both types of similarity graphs contribute to the final model performance, with $\mathcal{G}_1$ showing a more substantial impact.
The findings indicate that~\methodname~effectively captures subtle, continuous similarities between cells using the dual-channel cut-informed soft graph embedding module.
By minimizing Joint NCut, it effectively integrates cellular relationships across multiple similarity measures, robustly uncovering latent cellular relationships and enhancing clustering precision. 

\subsubsection{\textbf{Effectiveness of optimal transport}}
To illustrate our optimal transport-based clustering optimization module is better than the traditional approaches (e.g., DEC~\cite{xie2016unsupervised}, which adopts a clustering guided loss function to force the generated sample embeddings to have the minimum distortion against the pre-learned clustering center.), we design this experiment. 
Fig.~\ref{fig:ablation of all component} demonstrates scSGC's superiority over \emph{scSGC-w/o KL} across four datasets, highlighting the importance of optimal transport in the model. 
We further substituted DEC for our  optimal transport-based clustering optimization module within~\methodname~and compared the results. 
The results in Table~\ref{tab:ablation_opt} show that optimal transport significantly outperforms DEC across three key metrics. This indicates that optimal transport effectively maintains the balance of proximity guidance to pre-learned cluster centers, thus preventing degenerate solutions and achieving superior outcomes in the clustering tasks involving scRNA-seq data.

\subsection{\textbf{Parameter Sensitivity}}

\begin{figure*}[!t]
\centering
\subfloat[$\alpha$]{
\includegraphics[width=0.18\textwidth]{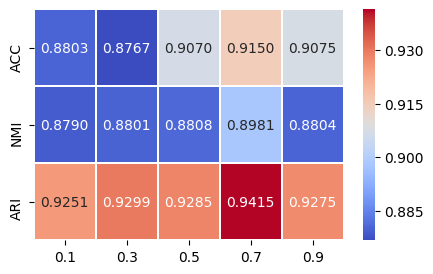}
}
\subfloat[$\beta$]{
\includegraphics[width=0.18\textwidth]{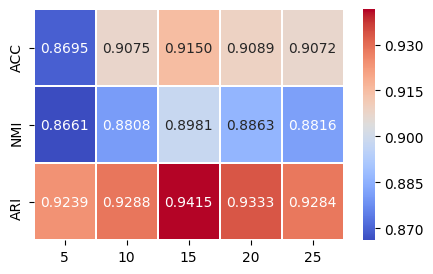}
}
\subfloat[$\gamma$]{
\includegraphics[width=0.18\textwidth]{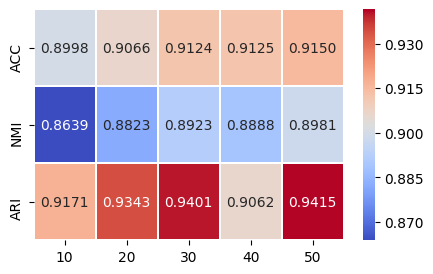}
}
\subfloat[$\mu$]{
\includegraphics[width=0.18\textwidth]{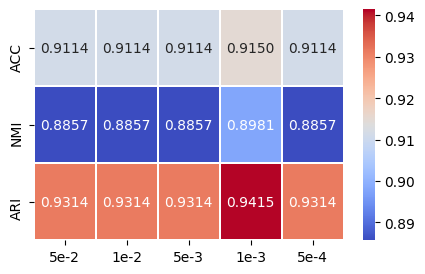}
}
\subfloat[$\lambda$]{
\includegraphics[width=0.18\textwidth]{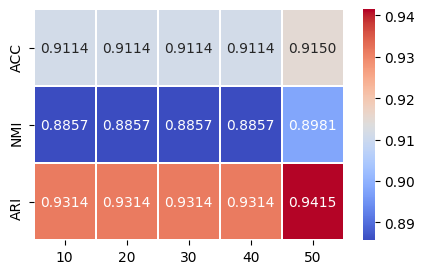}
}
\caption{Variation of~\methodname~performance on~\emph{Human Liver cells} dataset across different hyperparameters, including balance parameter $\alpha$, tuning parameters $\beta$, $\gamma$, $\mu$, and smoothness parameter $\lambda$.}
\label{fig:hyperparameters}
\end{figure*} 
Our work introduces five hyperparameters: $\alpha$, $\beta$, $\gamma$, $\mu$, and $\lambda$.
The specific roles of these hyperparameters are as follows:
\begin{itemize}
\item $\alpha$: A weight parameter that balances the contributions of the two graphs.
\item $\beta$: A regularization parameter used to adjust the impact of the regularization term.
\item $\gamma$ and $\mu$: Parameters that control the relative weights of the loss functions $\mathcal{L}_{ZINB}$ and $\mathcal{L}_{KL}$, respectively.
\item $\lambda$: A parameter that governs the trade-off between the accuracy of the assignment and the smoothness of the clustering in optimal transport.
\end{itemize}
We conducted a sensitivity analysis of the hyperparameters using the~\emph{Human Liver cells} dataset to investigate their impact on model performance. 
As shown in Fig.~\ref{fig:hyperparameters}, the experimental results demonstrate that~\methodname~maintains stable performance across a wide range of hyperparameter values. 
For instance, when $\mu$ varies between 0.1 and 0.9, the overall ACC value fluctuates around 0.91, indicating the robustness of this hyperparameter. 
Similarly, adjustments to other hyperparameters also lead to gradual changes in model performance, further confirming the model's stability with respect to hyperparameter settings.

To enhance the reproducibility of the experiments and the generalizability of the model, we recommend an initial set of hyperparameters: $\alpha=0.7$, $\beta=15$, $\gamma=50$, $\mu=1e-3$, and $\lambda=50$. 
These values serve as starting points for hyperparameter tuning, and adjusting them based on the specific characteristics of the data is necessary when applying~\methodname~to new datasets. Although our sensitivity analysis indicates that these settings generally yield near-optimal results across various datasets, fine-tuning the parameters to accommodate different datasets helps ensure the model's stability and optimal performance.
\subsection{\textbf{Computational Efficiency}}
\begin{figure}[t!]
    \centering
    \includegraphics[width=0.5\linewidth]{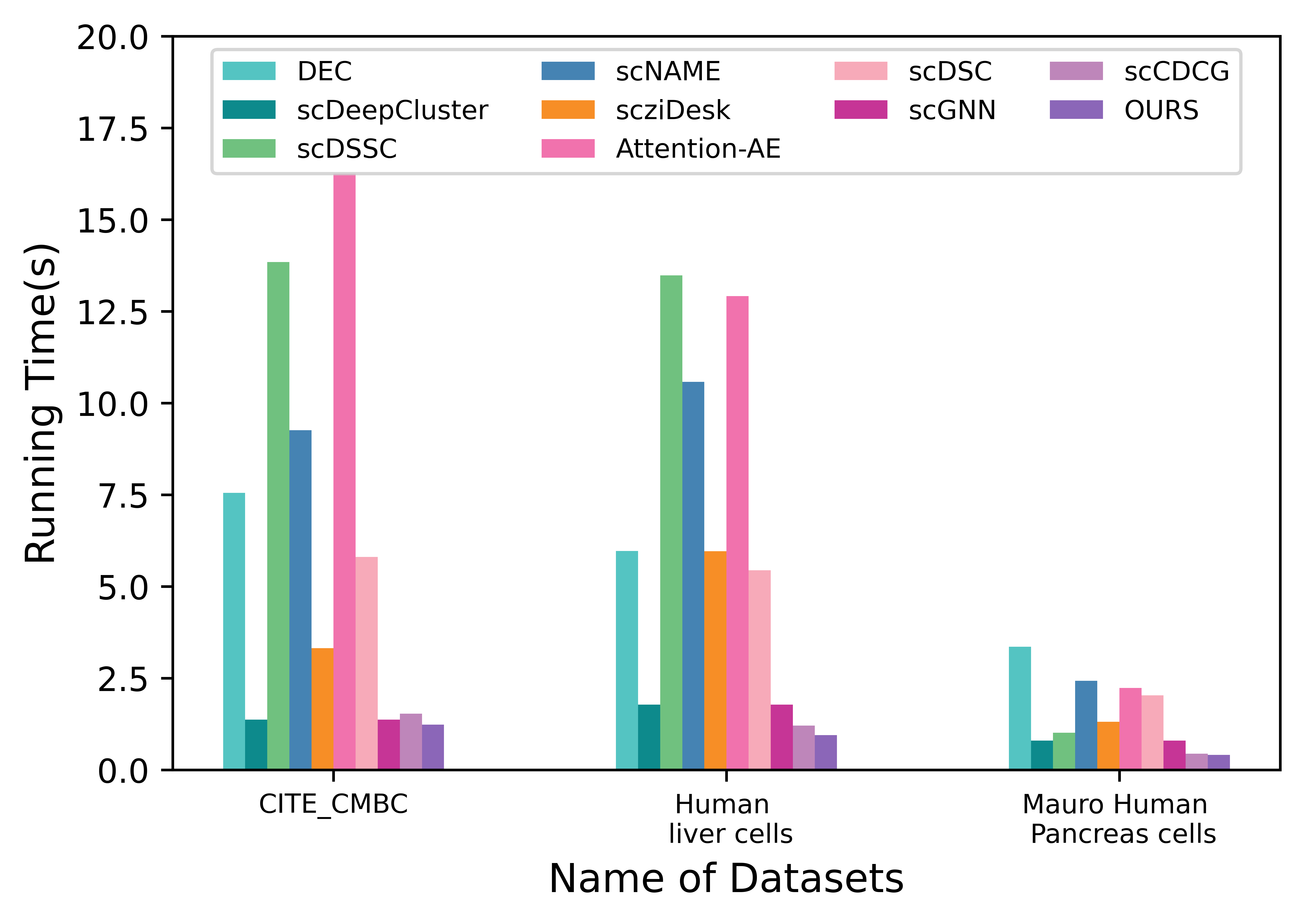}
    \caption{The running time of~\methodname~and nine baselines.}
    \label{fig:time_effectiveness}
\end{figure}

With the rapid advancement of scRNA-seq technology, the increasing size of datasets necessitates the development of highly efficient analysis methods. 
To systematically assess the computational efficiency of~\methodname, we performed runtime comparison experiments across three datasets, benchmarking against nine competing methods, including deep learning and graph neural network approaches.
All experiments were conducted on a server equipped with an EYPC 7742 processor, 1024GB RAM, and 8 Tesla A100 40GB GPUs. 
Fig.~\ref{fig:time_effectiveness} shows that~\methodname~consistently demonstrates superior computational efficiency across all datasets, while baseline models exhibit varying degrees of computational burden. 
~\methodname~maintains a relatively low and stable runtime across the tested datasets, and its scalability to substantially larger datasets (e.g., $>100k$ cells) requires further evaluation. 
Notably,~\methodname’s performance may vary depending on the computational hardware, and further optimization could be explored for environments with different resource constraints.

\subsection{Analysis of Simulated Data}
To comprehensively assess the robustness and generalization ability of the model across varying data conditions, we generated ten simulated datasets and conducted a systematic comparison between scSGC and its baseline method, scCDCG, which serves as the current benchmark with the best performance.
Specifically, we generated ten simulated scRNA-seq datasets using the R package Splatte~\cite{zappia2017splatter}, simulating both balanced datasets, where cell clusters are of approximately equal size, and imbalanced datasets, characterized by uneven cluster size distributions, under varying levels of dropout noise.
Each dataset consists of 3,000 cells, 2,500 genes, and six distinct cell clusters. To simulate varying levels of technical noise (i.e., dropout effects), we indirectly controlled the degree of sparsity by adjusting the mid parameter in Splatter (set to -12, -4, -2, -1.3, and -0.6, with shape = -1). These settings correspond to approximate dropout rates of 5\%, 10\%, 15\%, 20\%, and 25\%, respectively.

Notably, scSGC was developed based on the scCDCG framework, with hyperparameters fixed according to the configuration recommended in the "Parameter Sensitivity" section. 
These hyperparameters remained unchanged throughout the experiment to eliminate the potential influence of additional tuning on the final performance results.
As shown in Table~\ref{tab:simulated_comparison}, scSGC consistently outperformed scCDCG on all simulated datasets in terms of ACC, NMI, and ARI, thus validating the robustness and generalization capability of the proposed method under controlled conditions. A more detailed analysis of the results reveals the following key observations: (i) On balanced datasets, the overall performance of scSGC and scCDCG was generally better compared to imbalanced datasets, indicating that clustering methods are sensitive to class distribution.  (ii) scSGC demonstrated stable performance advantages even under challenging conditions, such as high dropout noise or severe class imbalance, reflecting its strong noise robustness and adaptability.  (iii) Particularly in imbalanced data scenarios, the performance improvement of scSGC over scCDCG was more pronounced, suggesting that the improvements made to the model effectively mitigated the negative impacts of data sparsity and class imbalance on clustering outcomes.

\begin{table*}[htbp]
\centering
\caption{Performance comparison between scSGC and scCDCG on ten simulated datasets (mean \textsubscript{$\scriptscriptstyle\pm$ std}).}
\tiny
\setlength{\tabcolsep}{1.5pt}  
\renewcommand{\arraystretch}{1.1}  
\begin{tabular}{llccccccccccc}
\toprule 
\multirow{2}{*}{\textbf{Method}} & \multirow{2}{*}{\textbf{Metric}} & \multicolumn{5}{c}{\textbf{Balanced data}} & \multicolumn{5}{c}{\textbf{Unbalanced data}} \\
\cmidrule(lr){3-7} \cmidrule(lr){8-12}
& & 5\% & 10\% & 15\% & 20\% & 25\% & 5\% & 10\% & 15\% & 20\% & 25\% \\
\midrule
\multirow{3}{*}{\textbf{scSGC}} 
& ACC & 82.7\textsubscript{$\scriptscriptstyle\pm$0.7} & 75.2\textsubscript{$\scriptscriptstyle\pm$4.4} & 82.6\textsubscript{$\scriptscriptstyle\pm$0.7} & 76.8\textsubscript{$\scriptscriptstyle\pm$0.7} & 73.8\textsubscript{$\scriptscriptstyle\pm$2.1} & 72.1\textsubscript{$\scriptscriptstyle\pm$2.1} & 84.4\textsubscript{$\scriptscriptstyle\pm$3.5} & 47.1\textsubscript{$\scriptscriptstyle\pm$0.8} & 74.0\textsubscript{$\scriptscriptstyle\pm$1.4} & 62.5\textsubscript{$\scriptscriptstyle\pm$4.3} \\
& NMI & 97.7\textsubscript{$\scriptscriptstyle\pm$1.9} & 74.2\textsubscript{$\scriptscriptstyle\pm$12.9} & 96.4\textsubscript{$\scriptscriptstyle\pm$0.7} & 76.4\textsubscript{$\scriptscriptstyle\pm$0.7} & 67.5\textsubscript{$\scriptscriptstyle\pm$2.1} & 60.9\textsubscript{$\scriptscriptstyle\pm$6.4} & 82.7\textsubscript{$\scriptscriptstyle\pm$1.3} & 27.5\textsubscript{$\scriptscriptstyle\pm$1.0} & 67.0\textsubscript{$\scriptscriptstyle\pm$1.4} & 60.0\textsubscript{$\scriptscriptstyle\pm$2.3} \\
& ARI & 98.2\textsubscript{$\scriptscriptstyle\pm$1.6} & 68.9\textsubscript{$\scriptscriptstyle\pm$12.0} & 97.1\textsubscript{$\scriptscriptstyle\pm$0.7} & 77.3\textsubscript{$\scriptscriptstyle\pm$0.7} & 66.1\textsubscript{$\scriptscriptstyle\pm$2.1} & 61.0\textsubscript{$\scriptscriptstyle\pm$2.1} & 85.0\textsubscript{$\scriptscriptstyle\pm$4.4} & 24.5\textsubscript{$\scriptscriptstyle\pm$4.4} & 61.6\textsubscript{$\scriptscriptstyle\pm$1.4} & 50.0\textsubscript{$\scriptscriptstyle\pm$6.0} \\
\midrule
\multirow{3}{*}{\textbf{scCDCG}} 
& ACC & 62.1\textsubscript{$\scriptscriptstyle\pm$2.8} & 61.1\textsubscript{$\scriptscriptstyle\pm$6.6} & 60.9\textsubscript{$\scriptscriptstyle\pm$1.0} & 64.6\textsubscript{$\scriptscriptstyle\pm$4.1} & 69.3\textsubscript{$\scriptscriptstyle\pm$8.0} & 67.0\textsubscript{$\scriptscriptstyle\pm$10.6} & 46.7\textsubscript{$\scriptscriptstyle\pm$2.1} & 40.9\textsubscript{$\scriptscriptstyle\pm$0.5} & 59.8\textsubscript{$\scriptscriptstyle\pm$5.8} & 58.3\textsubscript{$\scriptscriptstyle\pm$1.8} \\
& NMI & 62.7\textsubscript{$\scriptscriptstyle\pm$0.2} & 54.2\textsubscript{$\scriptscriptstyle\pm$4.9} & 50.7\textsubscript{$\scriptscriptstyle\pm$5.2} & 56.9\textsubscript{$\scriptscriptstyle\pm$9.6} & 67.3\textsubscript{$\scriptscriptstyle\pm$12.1} & 59.8\textsubscript{$\scriptscriptstyle\pm$16.6} & 37.1\textsubscript{$\scriptscriptstyle\pm$4.9} & 22.0\textsubscript{$\scriptscriptstyle\pm$0.3} & 50.0\textsubscript{$\scriptscriptstyle\pm$6.0} & 47.5\textsubscript{$\scriptscriptstyle\pm$5.8} \\
& ARI & 56.5\textsubscript{$\scriptscriptstyle\pm$2.7} & 46.7\textsubscript{$\scriptscriptstyle\pm$5.0} & 48.3\textsubscript{$\scriptscriptstyle\pm$5.0} & 52.8\textsubscript{$\scriptscriptstyle\pm$7.6} & 65.7\textsubscript{$\scriptscriptstyle\pm$10.6} & 56.5\textsubscript{$\scriptscriptstyle\pm$15.6} & 28.7\textsubscript{$\scriptscriptstyle\pm$3.2} & 19.1\textsubscript{$\scriptscriptstyle\pm$0.7} & 45.5\textsubscript{$\scriptscriptstyle\pm$8.0} & 41.3\textsubscript{$\scriptscriptstyle\pm$5.5} \\
\bottomrule
\end{tabular}
\label{tab:simulated_comparison}
\end{table*}

\section{\textbf{Conclusion}}

In conclusion, we propose~\methodname, an efficient and accurate framework for clustering single-cell RNA sequencing data. 
By integrating dual-channel soft graph representation learning with deep cut-informed techniques and incorporating ZINB-based feature autoencoder and optimal transport-driven clustering optimization,~\methodname~effectively addresses the critical challenges associated with traditional hard graph constructions, improving clustering accuracy while preserving biological relevance. 
This unified design enables~\methodname~to model complex, continuous relationships between cells while preserving biologically meaningful structures across diverse datasets.

The key contributions of this study are multifold. First, we formulate the clustering problem through a deep soft graph-cut perspective, enabling the development of a more flexible and accurate scRNA-seq clustering framework. Second, we propose a dual-channel, cut-informed soft graph learning module that effectively captures continuous similarities between cells. This design, in combination with the Joint NCut strategy, promotes consistency across learned graph structures and boosts downstream clustering performance. Third, we incorporate an optimal transport mechanism to refine clustering assignments, enabling minimal transport cost solutions that support both stability and biological coherence in cell population partitioning. Finally, we perform extensive benchmarking on ten diverse datasets, where~\methodname~demonstrates superior performance over eleven state-of-the-art clustering models across multiple criteria, including clustering accuracy, cell type annotation, and computational efficiency. These results highlight the robustness and generalizability of our framework in practical single-cell analysis settings.

Looking ahead, we intend to further strengthen~\methodname~ by incorporating advanced graph learning techniques and adapting the framework to accommodate other high-dimensional omics data. 
Further, we also plan to improve its scalability and efficiency to handle increasingly large and complex datasets.
These developments will likely open new avenues for understanding intricate cellular mechanisms and drive innovations in personalized medicine.

\section{Abbreviations}
scRNA-seq: single-cell RNA sequencing;\par
\noindent ZINB: zero-inflated negative binomial;\par
\noindent GNN: graph neural network;\par
\noindent SOTA: state-of-the-art;\par
\noindent KL: Kullback-Leibler;\par
\noindent ACC: Accuracy;\par
\noindent NMI: Normalized Mutual Information;\par
\noindent ARI: Adjusted Rand Index.

\section{Declarations}
\subsection{Ethics approval and consent to participate}
Not applicable. 
\subsection{Consent for publication}
Not applicable. 
\subsection{Availability of data and materials}
This manuscript does not report data generation, and all datasets used in this work are publicly available from open sources. 
All data and code required to reproduce the results presented in SGC will be made available on GitHub (\url{https://github.com/XPgogogo/scSGC}).
In addition, concerning the baseline methods, we have provided download links for the relevant code in the ``\texttt{readme.md}'' file on GitHub to facilitate access and verification. 

\subsection{Competing interests}
The authors declare that they have no competing interests.
\subsection{Funding}
This work is partially supported by the Strategic Priority Research Program of Chinese Academy of Sciences, Grant No.XDA0460104,  National Natural Science Foundation of China (No.92470204) and the Beijing Natural Science Foundation (No.4254089). 

\subsection{Authors' contributions}
PX conceptualized and designed the methodology, conducted all experiments, analyzed the results, and wrote the main manuscript text. PW contributed to parts of the manuscript writing and provided guidance on model design. Other authors participated in discussions on the methodology and offered guidance for the method design. All authors reviewed and approved the final manuscript. 
\subsection{Acknowledgements}
Not applicable. 


\bibliography{reference}

\end{document}